\pdfoutput=1

\documentclass[11pt, table]{article}

\usepackage[]{acl}
\usepackage{times}
\usepackage{latexsym}
\usepackage{multirow}
\usepackage{makecell}
\usepackage{graphicx,enumitem,amsmath,amssymb,tabularx,booktabs}
\usepackage{subcaption}
\usepackage{arydshln}

\makeatletter
\newcommand\notsotiny{\@setfontsize\notsotiny\@vipt\@viipt}
\makeatother

\newcolumntype{Y}{>{\centering\arraybackslash}X}

\usepackage[T1]{fontenc}

\usepackage[utf8]{inputenc}

\usepackage{microtype}

\usepackage{inconsolata}

\usepackage{graphicx}

\title{Are Economists Always More Introverted? \\ Analyzing Consistency in Persona-Assigned LLMs}
\author{Manon Reusens$^{1,2}$, Bart Baesens$^{1,3}$ and David Jurgens$^{4}$\\
         $^1$Research Centre for Information Systems Engineering (LIRIS), KU Leuven \\
         $^2$Department of Engineering Management, University of Antwerp \\
         $^3$Department of Decision Analytics and Risk, University of Southampton \\
         $^4$ School of Information, University of Michigan \\
         }%

\begin{document}
\maketitle
\begin{abstract}
Personalized Large Language Models (LLMs) are increasingly used in diverse applications, where they are assigned a specific persona—such as a happy high school teacher—to guide their responses. While prior research has examined how well LLMs adhere to predefined personas in writing style, a comprehensive analysis of consistency across different personas and task types is lacking. In this paper, we introduce a new standardized framework to analyze consistency in persona-assigned LLMs. We define consistency as the extent to which a model maintains coherent responses when assigned the same persona across different tasks and runs. Our framework evaluates personas across four different categories (happiness, occupation, personality, and political stance) spanning multiple task dimensions (survey writing, essay generation, social media post generation, single turn, and multi-turn conversations).  
Our findings reveal that consistency is influenced by multiple factors, including the assigned persona, stereotypes, and model design choices. Consistency also varies across tasks, increasing with more structured tasks and additional context.   All code is available on GitHub\footnote{\url{https://github.com/manon-reusens/persona_consistency}}.
\end{abstract}

\section{Introduction}
Personalized Large Language Models (LLMs) are increasingly deployed in applications where alignment with specific beliefs and values is essential, such as in high-stakes domains like healthcare and education~\cite{li-etal-2024-steerability,santurkar2023whose}. While prior research has examined the extent to which LLMs adhere to their assigned personas in terms of writing style~\cite{wang-etal-2024-incharacter,malik-etal-2024-empirical}, less attention has been given to the consistency of persona adherence across different types of tasks and prompting strategies~\cite{jiang-etal-2024-personallm}.
Moreover, it remains unclear how specifying certain persona attributes affects the consistency of \textit{other} characteristics. For example, does assigning an "economist" persona to an LLM ensure stable alignment across other characteristics, such as "extroversion"? Additionally, which persona categories lead to the most consistent behavior and does this depend on the task at hand? Addressing these questions is essential for understanding how LLM personas manifest across diverse contexts and for identifying unintended spillover effects—where defining an assigned persona might reinforce unintended other characteristics.  Recognizing both the intended and unintended traits associated with a persona is crucial for ensuring reliable and predictable model behavior, especially in high-stake environments.

Prior work shows that LLMs can reflect Big Five personality traits in structured tasks, and that larger models tend to do so more consistently than smaller ones~\cite{jiang-etal-2024-personallm,serapiogarcía2023personalitytraitslargelanguage}. However, persona consistency also extends beyond personality to traits like political orientation and social roles~\cite{rottger-etal-2024-political,shu-etal-2024-dont}.  Yet most evaluations rely on ad hoc methods, such as prompt perturbations or a narrow focus on personality-based personas, resulting in an incomplete picture of consistency. Interestingly, \citet{shu-etal-2024-dont} investigated whether explicitly assigning personas enhances response consistency. They found that while overall consistency decreased with persona assignment, responses became more consistent along dimensions relevant to the assigned persona. %

\begin{figure*}[h!]
    \centering
    \includegraphics[width=\linewidth]{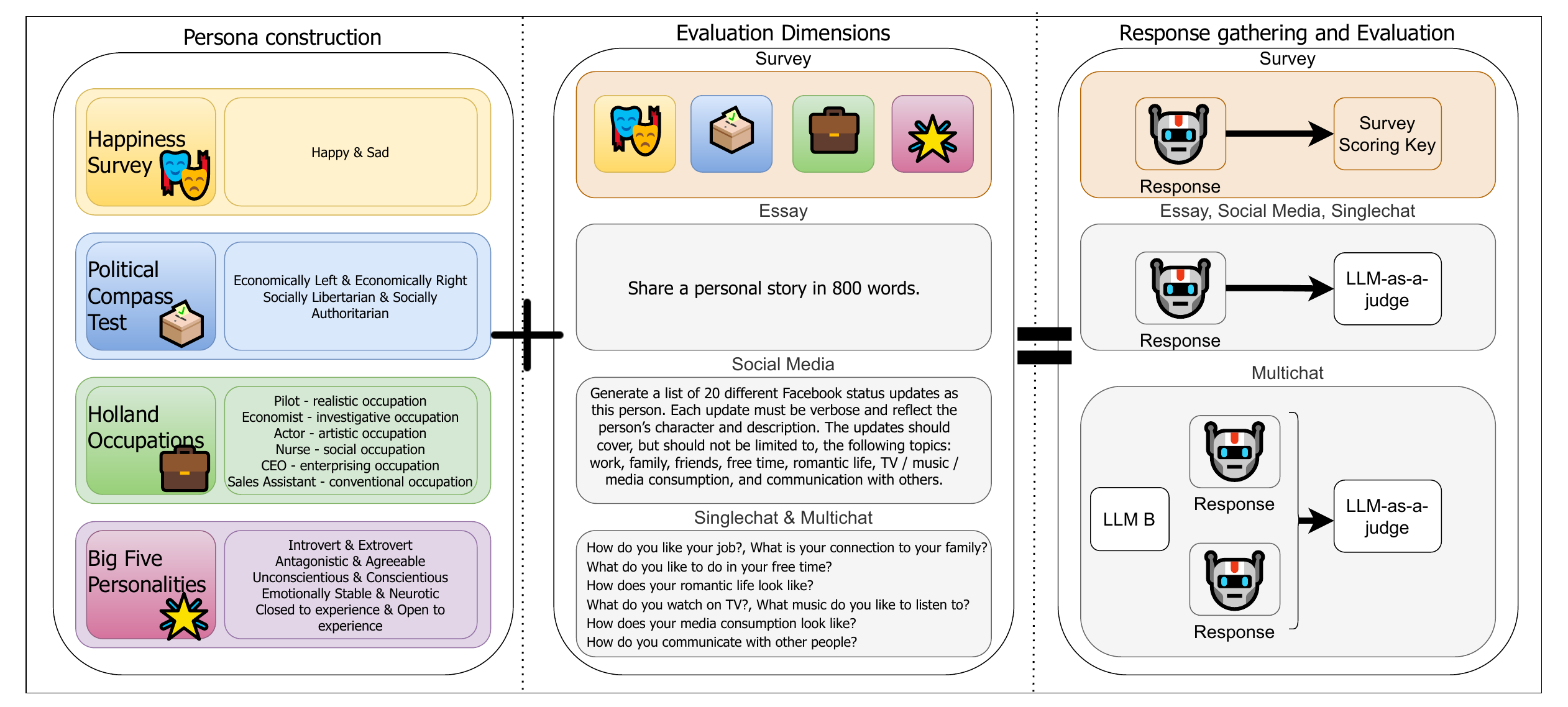}
    \caption{The overview of the full methodology. On the left, the persona construction is shown. From these four selected surveys, binary characteristics are selected serving as the base for the different personas. All combinations of these characteristics within a persona category are made, e.g. a character who is introverted, antagonistic, conscientious, neurotic, and open to experience. Surveys are evaluated using their respective scoring key. The other dimensions are evaluated using GPT4o as LLM-as-a-judge.}
    \label{fig:methodology}
    \vspace{-10pt}
\end{figure*}
 
 Given the recent calls for application-specific evaluations of LLM behavior~\cite{rottger-etal-2024-political,ouyang-etal-2023-shifted,zhaowildchat}, we propose a new standardized framework for analyzing persona consistency across a broader range of realistic tasks. Moving beyond prompt perturbations and personality-based personas, our approach provides a systematic analysis of consistency both within and across multiple persona categories and tasks.

More specifically, in this paper, we propose a standardized framework for multifaceted persona consistency analysis. We focus on four persona categories—happiness, occupation, personality, and political stance—selected for their relevance in persona-related literature, their variability in the scale of linguistic expression, and the availability of external survey instruments. We evaluate the consistency of persona-assigned LLMs across multiple evaluation dimensions, including survey answering, social media post generation, essay writing, single-question answering (singlechat), and multi-turn conversations (multichat).

We focus on two primary aspects of consistency:
\textbf{(1) Intra-persona consistency:} whether LLMs remain consistent within their assigned persona; and
\textbf{(2) Inter-persona consistency:} whether LLMs remain consistent across other persona categories than the one assigned.

We hypothesize that LLMs exhibit persona-dependent consistency effects: some personas (e.g., tied to specific professions) may induce stronger intra-persona consistency, while others (e.g., based on personality traits) lead to more partial or context-dependent patterns. We also explore potential spill-over effects—whether traits associated with one persona category influence outputs in another, possibly due to underlying social stereotypes.%

Our findings reveal that specifying a persona leads to high intra-persona consistency, with some persona categories (e.g., happiness and occupation) being more consistent than others (e.g., political stance). We also uncover spillover effects, where persona assignment reinforces inter-persona consistency driven by stereotypes and model defaults. Finally, we show that consistency is influenced by task dimensions: clearer tasks and additional context length improve consistency. %

\section{Framework and Metrics Development}

Here, we outline our framework and the evaluation metrics used in our experiments.

 \subsection{Persona construction} \label{sec:persona}
The left part of the consistency framework shown in Figure~\ref{fig:methodology} illustrates the persona construction. We selected the persona categories based on three key criteria: relevance in persona-related literature, variability in the scale of linguistic expression, and the availability of external survey instruments. Specifically, we focused on personality, professions, and political stance~\cite{malik-etal-2024-empirical,jiang-etal-2024-personallm,wang-etal-2024-incharacter}, which are well-studied categories in the persona-related literature. Additionally, we included a binary persona category (happy or sad) as a useful contrast, since emotional states tend to be more explicitly reflected in language, whereas categories like occupation may manifest more subtly. This range of personas allows us to examine varying degrees of consistency. Finally, for each identified persona category, we defined personas based on well-known surveys:
\begin{itemize}
    \item \textbf{Happiness:} The happiness personas were derived from the Happiness Survey developed by \citet{lyubomirsky1999measure}.
    \item \textbf{Political Stance:} These personas were based on the Political Compass Test ({\url{www.politicalcompass.org/test}})
    \item \textbf{Occupation:} Professional personas were determined using the survey outlined by~\citet{alma9929650410101488}. More specifically, we chose one occupation per occupation category defined by~\citet{alma9929650410101488}.
    \item \textbf{Personality:} These personas were assigned based on traits from the Big Five Inventory Test~\cite{john1999big}.
\end{itemize}
 
Additional information regarding the surveys used can be found in Appendix~\ref{app:surveys}. Based on the outcomes of the surveys, we constructed the different personas by making all possible combinations of the persona characteristics within a persona category, e.g. for the political category we include economically left and socially libertarian; economically right and socially libertarian; economically left and socially authoritarian; economically right and socially authoritarian. Additional information on the selection criteria for the persona categories are provided in Appendix~\ref{app:cat} and all personas are included in Appendix~\ref{app:personas}. This comprehensive approach ensures that our framework captures a wide range of realistic and nuanced persona scenarios. 

\subsection{Evaluation Dimension selection}
The surveys defined in Section~\ref{sec:persona}, not only guided the persona construction, but they also serve as one evaluation dimension for assessing the consistency of persona-assigned LLMs. In this evaluation dimension, LLMs are prompted to answer the survey questions individually in separate interactions. Additionally, we identified several other categories to analyze LLM personas: social media post generation, essay writing, single-question answering (singlechat), and multi-turn conversations (multichat). The prompts for these tasks were designed based on the methodologies outlined by \citet{serapiogarcía2023personalitytraitslargelanguage} and \citet{jiang-etal-2024-personallm}.  Based on established prompts for social media post generation, we distilled eight separate open-ended questions as initial prompts for both the singlechat and multichat evaluation dimensions. Multichat, in particular, was specifically designed for this study, building on the same initial prompts as singlechat. After the persona-assigned LLM generated its response, another LLM (LLaMA-3.2-1B) was introduced to engage with the reply. Next, the persona-assigned LLM received the full chat history and was prompted to respond once more. Consistency evaluation was conducted on both responses combined from the persona-assigned LLM. This process is also depicted on the right part of Figure~\ref{fig:methodology}. All tasks were carefully selected to align with the call for application-specific evaluations~\cite{rottger-etal-2024-political,ouyang-etal-2023-shifted,zhaowildchat}, ensuring our analysis captures the real-world relevance and practical adaptability of persona-assigned LLMs.

\subsection{Scoring Key}
Additional information on the scoring keys is provided below. A more detailed explanation, including the prompts and specific survey scoring mechanisms, can be found in Appendix~\ref{app:eval}.
\paragraph{Survey Dimension.}
The survey dimension is evaluated using its respective scoring methodology. To ensure consistency in our analysis, final results are simplified into binary categories for all surveys except the occupational one, where the primary relevant occupational category is selected. For the happiness survey, responses are categorized as happy or sad. In the political compass test, the persona-assigned LLM’s outputs are analyzed to determine the corresponding quadrant, with the final outcome identified across both the economic and social axes. In the personality survey, the outputs are assessed within the framework of the Big Five traits and classified into binary categories per trait. 

\paragraph{Open Response Dimensions.}
For the other dimensions, we use an LLM-as-a-judge, GPT4o, to evaluate the final outcomes, determining the persona's alignment with binary characteristics, or for occupations one of the six different classes. The evaluation process is consistent across all dimensions, with the LLM assessing all characteristics across all responses.
Detailed information on the used prompt is provided in Appendix~\ref{app:llmjudge}. The model also provides a confidence score on a four-point Likert scale per choice. A neutral choice was identified when the model's confidence score was 1 or 2.  To validate the reliability of the LLM judgments, we manually annotated 100 random examples across all evaluation categories from our dataset, with 25 samples per assigned persona category. This resulted in 50 sentences from singlechat, 42 from multichat, 4 from essay, and 4 from socialmedia. We acknowledge the distribution skews toward singlechat and multichat, which are notoriously challenging for consistency evaluations due to conversational variability and nuanced phrasing. We found a Cohen's $\kappa$ of 0.68 with these final LLM results supporting the reliability of the LLM-generated judgments. Following ~\citet{8d20e0b8-89d8-3d65-bcf5-8c19d56ec4ab}, this score lies within the substantial agreement range. Given that most of the validated samples represent the hardest categories for evaluation, this level of agreement is, in our opinion, a reasonably strong indication that the LLM’s judgments align with human intuitions under challenging conditions. Additionally, this task is already rather difficult for people. For example, \citet{preoctiuc2015studying} show how inferring subjective characteristics about users from socialmedia posts such as optimism, leads to cohen’s kappa ranges between 0.3 and 0.7.

\subsection{Consistency scores} \label{sec:cons_scores}
\paragraph{Entropy.}
To measure consistency, we use Shannon entropy~\citep{shannon1948mathematical}, a metric that captures the uncertainty in the distribution of predicted labels across responses. It applies to both binary and multiclass persona traits and reflects full distributional patterns rather than just majority labels. Crucially, entropy allows us to compare consistency across models and persona categories without relying on arbitrary thresholds. Additional information about the choice of the metric is included in Appendix~\ref{app:entropy}

An entropy score quantifies how focused or scattered the model's responses are. For example, if a model consistently outputs "happy" for a happiness persona across multiple prompts, the entropy is low, indicating high consistency. If the responses are split between "happy" and "sad", the entropy is higher, signaling inconsistency.

We compute entropy for each system prompt $s$ within an evaluation category $e$, persona category $p$, and dimension $d$ as:
\begin{equation}
    entropy_{s_e,p,d}=- \frac{\sum_{x \in X} P(x)*\log (P(x))}{\log(|X|) }
\end{equation}
Here, $X$ is the set of possible characteristics (e.g., for happiness: {happy, sad}), and $P(x)$ is the proportion of responses labeled with characteristic $x$. All persona categories are binary except for occupation, which includes six possible labels. 

The probability scores per characteristic are calculated using the labels from the LLM-as-a-judge for the different responses. We added the neutral category as a random prediction to every option in the underlying characteristic, because a neutral response indicates a lack of alignment with the intended characteristic. Since consistency requires a persona to manifest in a discernible way, we also treat neutral predictions as inconsistent.

Finally, we compute a single entropy score for each persona and evaluation category by averaging across all system prompts and dimensions:
\begin{equation}
    entropy_{p,e}=\frac{1}{|D|}\sum_{d \in D}\frac{1}{|S|}\sum_{s \in S} entropy_{s_e,p,d} 
\end{equation}
 
\paragraph{Characteristic-specific consistency.}
The main disadvantage of the entropy metric is that it does not show which attribute the LLM consistently outputs. Hence, we also examine the average scores per persona characteristic. For binary characteristics, we use a continuous scale from 0 to 1, where both endpoints represent distinct, persona-aligned responses. A score of 0.5 indicates a lack of alignment with the underlying characteristic, stemming from inconsistency or neutrality. Higher consistency is found when scores are closer to 0 or 1. %

For the occupation category, we determine the most frequently assigned occupation category and identify the intensity score as the probability of occurrence. A perfectly consistent model receives an intensity score of 1, while a randomly distributed model is expected to score 1/6.

\section{Consistency Analysis}
In this section, we present the research questions, the experimental setup, and the results.
\subsection{Research Questions}
In this study, we examine the consistency of persona-assigned LLMs and spillover effects across different persona categories. Specifically, we address the following research questions:
\begin{enumerate}
    \item \textbf{RQ1 Intra-persona consistency:} How does assigning a persona to an LLM result in differences in intra-persona consistency across various persona categories?
    \item \textbf{RQ2 Spillover effects:} Does assigning a persona to an LLM lead to spillover effects in other, unspecified persona categories?
    \item \textbf{RQ3 Cross-dimensional consistency:} How does consistency vary across different response dimensions, particularly with the inclusion of the multichat dimension?

\end{enumerate}

\subsection{Experimental set-up}
We analyze the consistency over 5 runs across 5 models from 3 different model families: Qwen-2.5 32B, Ministral-8B, Llama-3.2 3B, Llama-3.1 8B, and Llama-3.3 70B. Additional information on the checkpoints used is included in Appendix~\ref{app:model}
We analyze the entropy per model and per evaluation and persona category. To gather insights into the chosen labels per characteristic, we examine the characteristic-specific consistency. Next, we analyze the overall consistency making cross-dimension comparisons. 

\subsection{Results}

\begin{table*}[ht!]
\centering
    \begin{subtable}[t]{0.48\linewidth}
    \centering 
    \resizebox{\linewidth}{!}{%
    \begin{tabular}{lcccc}
    \toprule
    \textit{Evaluation Categories} & \multicolumn{4}{c}{\textit{Persona Categories}} \\
    \cmidrule(lr){2-5}
         & Happiness & Occupation & Personality & Political  \\
         \midrule
        Happiness   & \cellcolor{orange!25}$0.28 \pm 0.26$  & \cellcolor{orange!25}$0.25 \pm 0.41$  & \cellcolor{green!25}$0.15 \pm 0.09$  & \cellcolor{green!25}$0.20 \pm 0.44$  \\
        Occupation  & \cellcolor{red!25}$0.55 \pm 0.38$   & \cellcolor{green!25}$0.21 \pm 0.21$  & \cellcolor{red!25}$0.50 \pm 0.33$  & \cellcolor{red!25}$0.50 \pm 0.33$  \\
        Personality & \cellcolor{green!25}$0.19 \pm 0.13$ & \cellcolor{green!25}$0.16 \pm 0.11$  & \cellcolor{orange!25}$0.25 \pm 0.15$  & \cellcolor{green!25}$0.22 \pm 0.11$  \\
        Political   & \cellcolor{red!25}$0.80 \pm 0.45$   & \cellcolor{red!25}$0.77 \pm 0.43$  & \cellcolor{red!25}$0.75 \pm 0.40$  & \cellcolor{orange!25}$0.39 \pm 0.46$  \\
    \bottomrule
    
   \end{tabular}%
    }
    \vspace{-5pt}
    \caption{Entropy scores for Qwen-2.5 32B. }
    \label{tab:entropy_qwen}
    \end{subtable}
    \hfill
    \begin{subtable}[t]{0.48\linewidth}
        \resizebox{\linewidth}{!}{%
    \begin{tabular}{lcccc}
    \toprule
    \textit{Evaluation Categories} & \multicolumn{4}{c}{\textit{Persona Categories}} \\
    \cmidrule(lr){2-5}
         & Happiness & Occupation & Personality & Political  \\
         \midrule
        Happiness   & \cellcolor{orange!25}$0.26 \pm 0.25$  & \cellcolor{orange!25}$0.28 \pm 0.38$  & \cellcolor{orange!25}$0.38 \pm 0.17$  & \cellcolor{orange!25}$0.40 \pm 0.35$  \\
        Occupation  & \cellcolor{red!25}$0.76 \pm 0.18$   & \cellcolor{orange!25}$0.40 \pm 0.32$  & \cellcolor{red!25}$0.68 \pm 0.16$  & \cellcolor{red!25}$0.57 \pm 0.21$  \\
        Personality & \cellcolor{orange!25}$0.42 \pm 0.15$ & \cellcolor{green!25}$0.24 \pm 0.15$  & \cellcolor{orange!25}$0.39 \pm 0.11$  & \cellcolor{orange!25}$0.37 \pm 0.08$  \\
        Political   & \cellcolor{red!25}$0.91 \pm 0.17$   & \cellcolor{red!25}$0.93 \pm 0.07$  & \cellcolor{red!25}$0.85 \pm 0.15$  & \cellcolor{red!25}$0.51 \pm 0.35$  \\
    \bottomrule
   \end{tabular}%
    }
    \vspace{-5pt}
    \caption{Entropy scores for Ministral-8B. }
    \label{tab:entropy_ministral}
    \end{subtable}
    \vspace{5pt}
    \vfill
    \begin{subtable}[t]{0.48\linewidth}
    \centering
    \resizebox{\linewidth}{!}{%
    \begin{tabular}{lcccc}
    \toprule
    \textit{Evaluation Categories} & \multicolumn{4}{c}{\textit{Persona Categories}} \\
    \cmidrule(lr){2-5}
         & Happiness & Occupation & Personality & Political  \\
         \midrule
        Happiness   & \cellcolor{green!25}$0.00 \pm 0.00$  & \cellcolor{orange!25}$0.26 \pm 0.28$  & \cellcolor{red!25}$0.52 \pm 0.11$  & \cellcolor{red!25}$0.65 \pm 0.17$  \\
        Occupation  & \cellcolor{red!25}$0.70 \pm 0.13$   & \cellcolor{orange!25}$0.38 \pm 0.26$  & \cellcolor{red!25}$0.64 \pm 0.21$  & \cellcolor{red!25}$0.62 \pm 0.20$  \\
        Personality & \cellcolor{orange!25}$0.31 \pm 0.18$ & \cellcolor{orange!25}$0.31 \pm 0.13$  & \cellcolor{orange!25}$0.43 \pm 0.11$  & \cellcolor{orange!25}$0.47 \pm 0.17$  \\
        Political   & \cellcolor{red!25}$0.86 \pm 0.22$   & \cellcolor{red!25}$0.81 \pm 0.25$  & \cellcolor{red!25}$0.89 \pm 0.13$  & \cellcolor{red!25}$0.75 \pm 0.13$  \\
    \bottomrule
   \end{tabular}%
    }
    \vspace{-5pt}
    \caption{Entropy scores for Llama-3.2-3B.}
    \label{tab:entropy_llama3B}
    \end{subtable}
    \hfill
    \begin{subtable}[t]{0.48\linewidth}
    \centering
    \resizebox{\linewidth}{!}{%
    \begin{tabular}{lcccc}
    \toprule
    \textit{Evaluation Categories} & \multicolumn{4}{c}{\textit{Persona Categories}} \\
    \cmidrule(lr){2-5}
         & Happiness & Occupation & Personality & Political  \\
         \midrule
        Happiness   & \cellcolor{green!25}$0.07 \pm 0.16$  & \cellcolor{green!25}$0.22 \pm 0.14$  & \cellcolor{orange!25}$0.42 \pm 0.12$  & \cellcolor{orange!25}$0.43 \pm 0.10$  \\
        Occupation  & \cellcolor{red!25}$0.67 \pm 0.29$   & \cellcolor{green!25}$0.23 \pm 0.17$  & \cellcolor{red!25}$0.63 \pm 0.24$  & \cellcolor{red!25}$0.57 \pm 0.21$  \\
        Personality & \cellcolor{orange!25}$0.31 \pm 0.15$ & \cellcolor{green!25}$0.24 \pm 0.07$  & \cellcolor{orange!25}$0.36 \pm 0.13$  & \cellcolor{orange!25}$0.41 \pm 0.10$  \\
        Political   & \cellcolor{red!25}$0.83 \pm 0.33$   & \cellcolor{red!25}$0.76 \pm 0.38$  & \cellcolor{red!25}$0.79 \pm 0.30$  & \cellcolor{orange!25}$0.35 \pm 0.30$  \\
    \bottomrule
   \end{tabular}%
    }
    \vspace{-5pt}
    \caption{Entropy scores for Llama-3.1-8B.}
    \label{tab:entropy_llama8B}
\end{subtable}
\vspace{5pt}
\vfill
\begin{subtable}[t]{0.48\linewidth}
\centering
    \resizebox{\linewidth}{!}{%
    \begin{tabular}{lcccc}
    \toprule
    \textit{Evaluation Categories} & \multicolumn{4}{c}{\textit{Persona Categories}} \\
    \cmidrule(lr){2-5}
         & Happiness & Occupation & Personality & Political  \\
         \midrule
        Happiness   & \cellcolor{green!25}$0.07 \pm 0.15$  & \cellcolor{green!25}$0.08 \pm 0.09$  & \cellcolor{orange!25}$0.30 \pm 0.18$  & \cellcolor{green!25}$0.20 \pm 0.16$  \\
        Occupation  & \cellcolor{red!25}$0.51 \pm 0.38$   & \cellcolor{green!25}$0.22 \pm 0.18$  & \cellcolor{red!25}$0.55 \pm 0.33$  & \cellcolor{orange!25}$0.46 \pm 0.35$  \\
        Personality & \cellcolor{green!25}$0.23 \pm 0.14$ & \cellcolor{green!25}$0.17 \pm 0.10$  & \cellcolor{orange!25}$0.27 \pm 0.18$  & \cellcolor{green!25}$0.23 \pm 0.14$  \\
        Political   & \cellcolor{red!25}$0.79 \pm 0.44$   & \cellcolor{red!25}$0.76 \pm 0.43$  & \cellcolor{red!25}$0.71 \pm 0.40$  & \cellcolor{orange!25}$0.36 \pm 0.30$  \\
    \bottomrule
   \end{tabular}%
    }
    \vspace{-5pt}
    \caption{Entropy scores for Llama-3.3 70B. }
    \label{tab:entropy_llama70B}
\end{subtable}
\caption{The tables show large entropy differences between the models, indicating differences in consistency levels. Both strong within-category consistency (diagonal) and occasional spill-over consistency (off-diagonal) are found, where lower is more consistent. Scores $<$0.25 are colored green, 0.25$<$0.5 in orange, and $>$0.5 in red. The columns represent the different assigned persona categories. The rows represent the evaluation categories. The standard deviation is computed over the entropy scores over different dimensions per evaluation-persona category pair.}
\label{tab:entropy_all}
\end{table*}

\begin{figure}[t!]
  \centering
  \begin{subfigure}[b]{\linewidth}
    \centering
    \includegraphics[width=\linewidth]{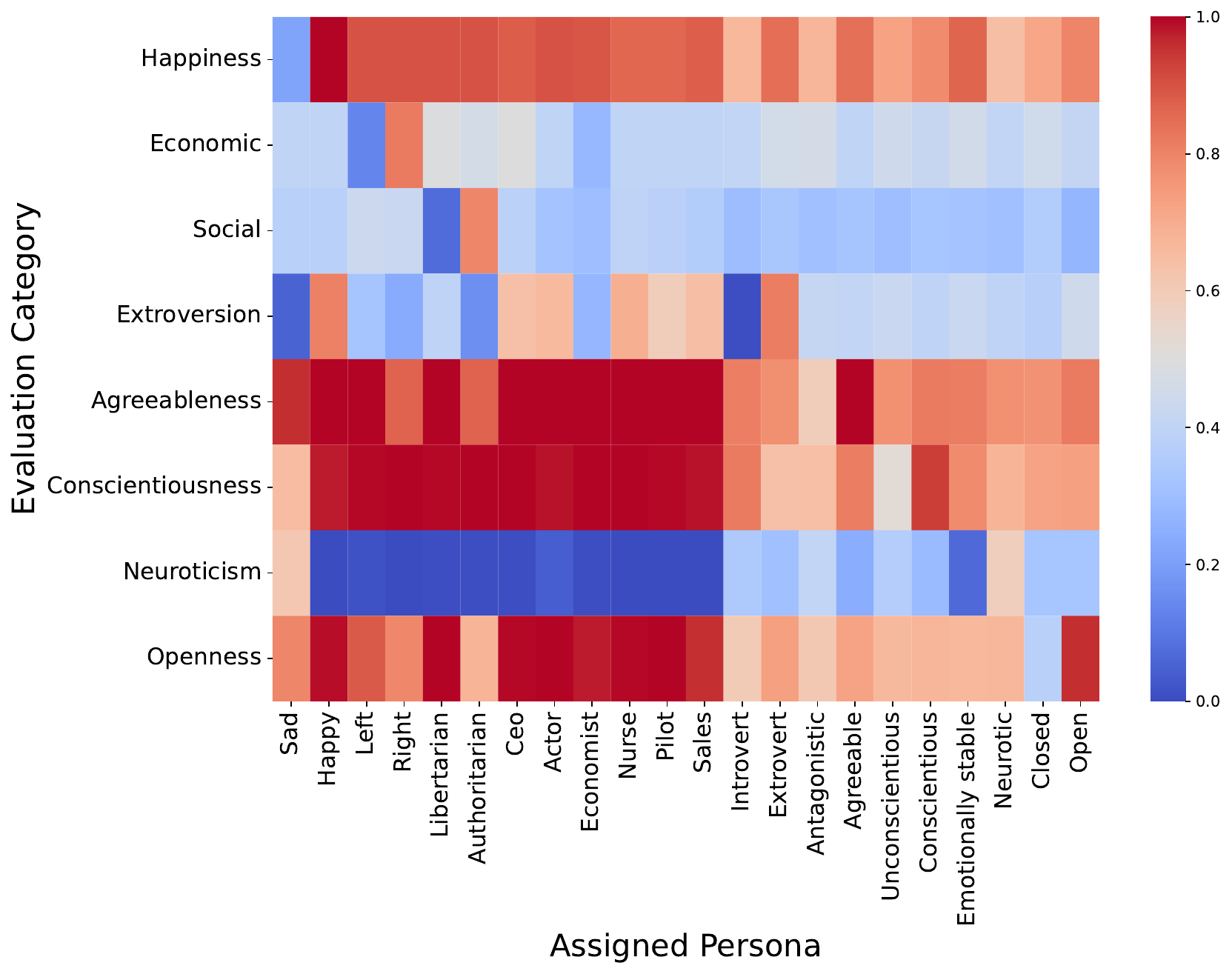}
    \caption{Heatmap providing the characteristic-specific consistency for all evaluation categories except occupation. A score of 1 favors the category name, 0 favors its opposite (e.g., agreeableness vs. antagonistic), and 0.5 indicates inconsistency.}
    \label{fig:heatmap}
  \end{subfigure}
  \vspace{10pt} %
  \begin{subfigure}[b]{\linewidth}
    \hfill \includegraphics[width=0.9\linewidth]{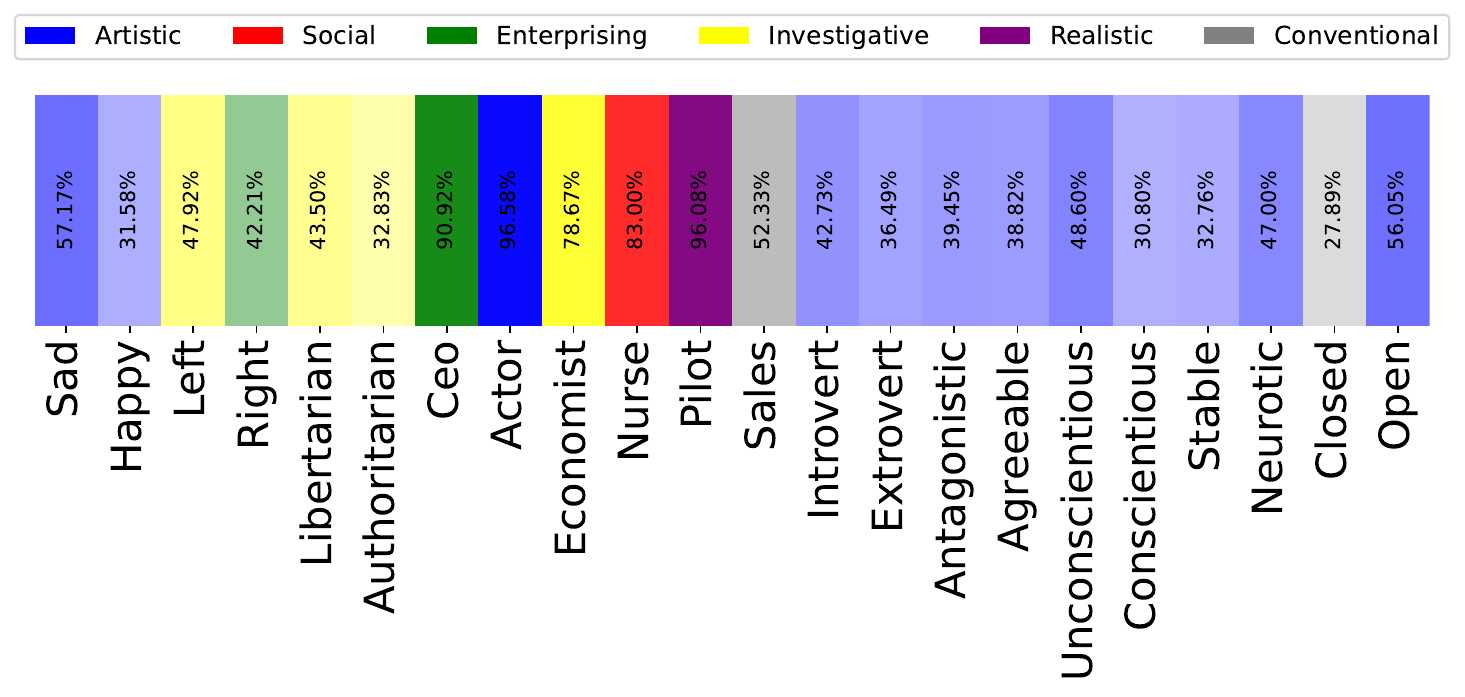}
    \caption{This Figure shows the most dominant occupation category per persona across experiments. Color intensity represents the consistency score per label.}
    \label{fig:occupation_heatmap}
    \vspace{-10pt}
  \end{subfigure}
  \caption{Qwen-2.5 32B generally follows the instructions and shows spill-over effects, i.e. stereotypes and default values. Columns denote personas, and rows indicate evaluation categories. Multi-component personas (e.g., political stance) are grouped per component and averaged over all personas containing that component. The other models are shown in Appendix~\ref{app:char}.  }
  \label{fig:overall_heatmap}
\vspace{-16pt}
\end{figure}

\paragraph{Intra-persona consistency is high within each category, but notable differences emerge across categories (RQ1).} As shown in Table~\ref{tab:entropy_all}, the diagonal values indicate relatively high consistency, meaning that when a persona is assigned, the model tends to generate responses that consistently express that specific persona across different output formats and prompts. However, the degree of consistency varies across persona categories. While happiness and occupation personas are more consistently expressed, personality and political personas exhibit lower intra-persona consistency. For the political category, we observe a high standard deviation, indicating substantial variability in consistency across dimensions. This is largely due to certain tasks, such as singlechat, where expressing a consistent political stance is more challenging in general. Manual analysis also confirms this finding, highlighting the difficulty of a model to express the political opinion when being asked certain questions. Here, the inconsistency thus stems from a lack of expression of the underlying persona. Similarly, personality-based personas show lower intra-persona consistency, which we investigate further in Figure~\ref{fig:overall_heatmap}. This figure shows the results for Qwen-2.5 32B (other models are shown in Appendix~\ref{app:char}). We chose  Qwen as it provides representative results for all models with an average correlation of 0.66 with the other models. Examining Figure~\ref{fig:overall_heatmap}, we find that the LLM generally follows our instructions, e.g., the happy persona is more happy (1), while the sad persona is more sad (0).  Likewise assigned occupations are clearly reflected in the output. However, certain personality traits, such as low conscientiousness, antagonism, neuroticism, and, in some models, low openness to experience, are more difficult for the LLM to express consistently, resulting in greater variability in responses within that personality category. Similarly,~\citet{10.1093/pnasnexus/pgae533} show that models skew responses to socially desirable answers for the Big Five Personality test when they infer that they are evaluated. We demonstrate that social desirability bias also appears in other evaluation dimensions. This social desirability tendency explains the model's difficulty to adhere to the unconscientious and antagonistic personas.

\paragraph{Spill-over effects vary across evaluation categories (RQ2). }
The off-diagonal elements in the subtables of Table~\ref{tab:entropy_all} reveal that most spill-over consistency effects occur across two evaluation categories: happiness and personality. For example, the off-diagonal entropy scores for Llama-3.3 70B are lower for the happiness and personality categories compared to occupation and political stance. This suggests that the model is more consistent across these two categories when they are not the assigned persona. Interestingly, this pattern differs from the intra-persona results, where happiness and occupation showed the strongest consistency when they were the assigned persona. The occupation and political categories are harder for the models to express, as not adding those personas results in responses without any occupational information or political stance. Manual analysis reveals how these personas are less frequently and explicitly expressed, especially in conversational settings. For instance, political beliefs rarely surface in responses to questions like \textit{"What are your music preferences?"}. Similarly, occupation-related information rarely appears unless explicitly prompted, though it may occasionally surface in essay-style or social media posts. The other two categories show spill-over effects. Additional manual analysis reveals that happiness and personality directly influence linguistic style: models default to a positive tone unless instructed otherwise, making happiness more overt. Similarly, personality traits, like extroversion, shape response style, amplifying spillover effects.  

\paragraph{Spill-over effects are due to two main factors: stereotypical associations with the assigned persona and default personas when a characteristic is not explicitly assigned (RQ2).} Figure~\ref{fig:overall_heatmap} shows how a sad persona is portrayed as more introverted, less conscientious, and less open than a happy persona. The economically right-winged and socially authoritarian personas are both less agreeable than their counterparts. All occupation personas are presented as extroverts, except the economist, who is more introverted. These observations reinforce prior findings that persona-assigned LLMs are susceptible to stereotypes~\cite{guptabias}. We show that these stereotypes also appear across general text-generation tasks. Furthermore, the model tends to answer in a happy, conscientious, agreeable, and open manner unless otherwise instructed or influenced by a stereotype. This is reflected in the heatmaps: the rows corresponding to the happiness, conscientiousness, and agreeableness evaluation categories generally lean toward a value of 1, indicating strong alignment. In contrast, neuroticism tends to lean toward 0, suggesting lower identification with that trait. This again illustrates the social desirability bias in the models ~\cite{10.1093/pnasnexus/pgae533}. Additionally, we show how personas can partially counteract this bias, e.g. the high consistency scores for neurotic and sad personas. However, this effect is not universal, as evidenced by the lower intra-persona consistency for antagonistic and unconscientious personas. Furthermore, most models tend to be slightly economically left and socially libertarian, though this varies by model and sometimes leans toward inconsistency. These default personas reveal the LLM's ideological stances, shaped by training data and choices from model developers, called design choices~\cite{buyl2025largelanguagemodelsreflect,cambo2022model}.

\paragraph{Consistency is higher in more structured tasks like survey answering. For open-ended question answering tasks, providing additional context through multi-turn interactions (multichat) improves consistency (RQ3).}
\begin{figure}[t!]
    \centering
    \includegraphics[width=\linewidth]{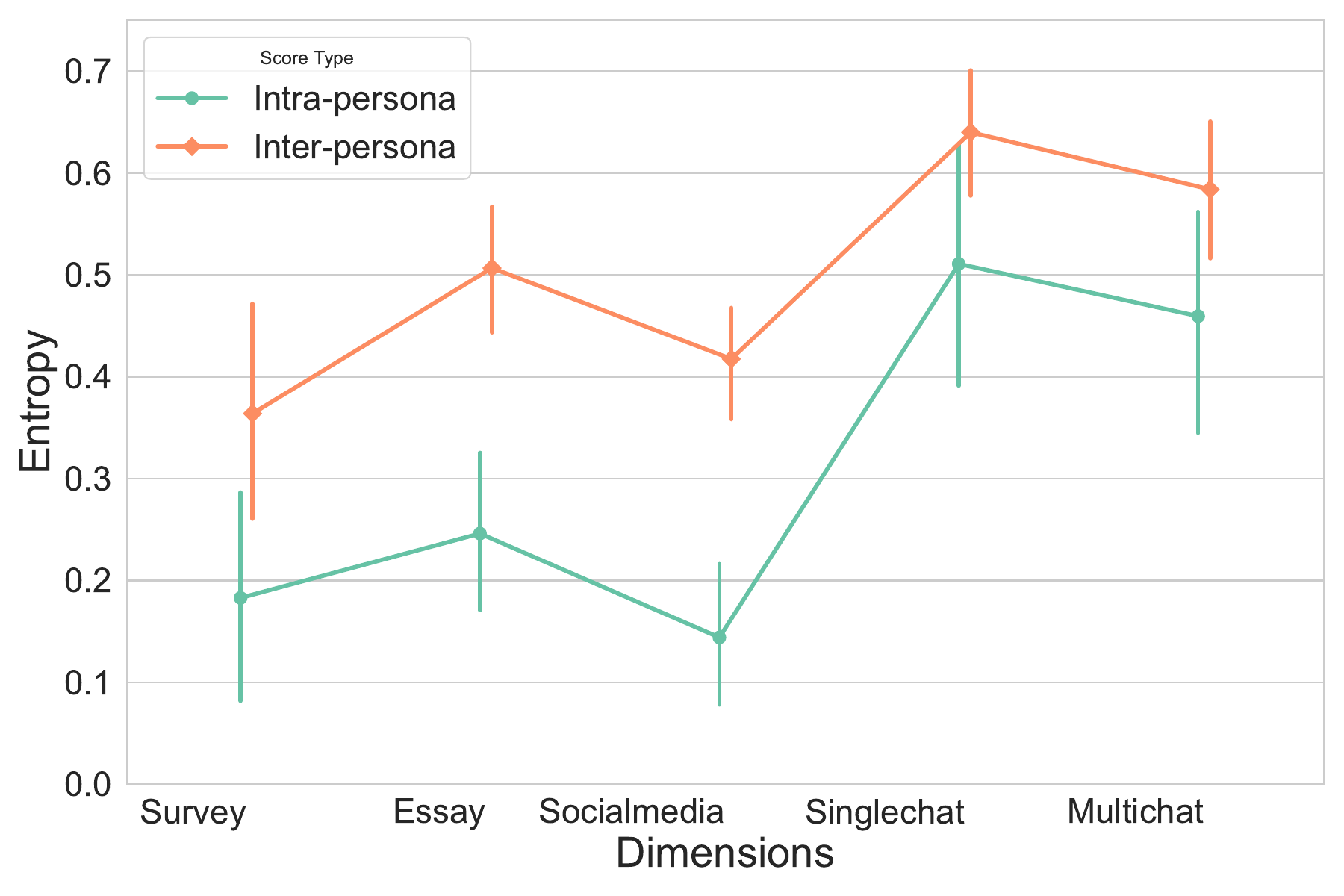}
    \caption{The average intra-persona and inter-persona entropy across all models per dimension reveals large differences in entropy between different dimensions. Error bars represent 95\% confidence intervals estimated via bootstrapping, using nonparametric resampling to approximate the uncertainty around the mean.}
    \label{fig:pointplot_dimensions}
    \vspace{-15pt}
\end{figure}

Figure~\ref{fig:pointplot_dimensions} shows that the survey, essay, and social media dimensions have the lowest intra- and inter-persona entropy, indicating the highest consistency across these tasks. However, consistency scores vary notably across dimensions, with mostly low correlations between them. This highlights the importance of considering all three dimensions to fully capture model behavior. Of these, only the inter-persona consistency between essay and social media is strongly correlated. Their low entropy scores could be attributed to the clarity of the task, where models can easily express their assigned persona. As tasks become less straightforward—such as answering open-ended questions about music preferences (singlechat and multichat)—models generally show a decrease in both intra- and inter-persona consistency. Moreover, the difference between intra- and inter-persona entropy becomes smaller. These results suggest that as tasks require less explicit persona expression, models may struggle to express distinct persona characteristics. Depending on the application, this variability may be advantageous, i.e. in creative or open-ended generation tasks where diversity is desirable. However, in more controlled settings that require reliable persona adherence, such unpredictability can be a drawback, as the model’s outputs may deviate from the intended persona or produce inconsistent behavior. Interestingly, the complexity of multichat scenarios compared to singlechat does not appear to hinder consistency. Contrarily, consistency increases slightly as follow-up responses allow models to provide more information, expressing the assigned persona more clearly. Nevertheless, consistency scores between singlechat and multichat are highly correlated. In addition to the main results, supplementary statistical analyses revealed no significant difference between inter- and intra-persona consistency in the singlechat and multichat dimensions. However, we observed statistically significant differences between the evaluation dimensions of survey, essay, and social media on the one hand, and singlechat (for both intra- and inter-persona consistency) as well as multichat (for intra-persona consistency) on the other. Concerning inter-persona consistency specifically, only the social media post generation task significantly outperforms the inter-persona consistency observed in the multichat evaluation dimension. Detailed statistical analysis is reported in Appendix~\ref{stat_analysis}.

\section{Discussion}
Our framework offers a multi-dimensional perspective on the consistency of persona-assigned LLMs. Our results show that the balance between intra-persona and inter-persona consistency varies depending on the evaluation dimension. Persona-based models are more consistent for attributes explicitly defined in their persona than for unspecified persona attributes, indicating a weaker spillover effect.  However, in singlechat and multichat settings, we find no statistical difference between inter- and intra-persona consistency, implying that adding a persona does not significantly increase consistency compared to non-assigned attributes in more realistic tasks.  This finding urges caution when deploying persona-assigned LLMs in practical applications, as persona adherence may be less consistent in conversational contexts than in more structured tasks like surveys or essays. Finally, we show that longer context, i.e. comparing multichat to singlechat, allows models to express their persona more clearly leading to higher consistency.

We identify three main factors influencing consistency across different characteristics. 
\paragraph{The assigned persona:} Models generally adhere well to persona-specific instructions, with the degree of adherence varying across categories, models and dimensions. Our paper offers a valuable comparison on consistency levels across different persona categories, highlighting how some categories (e.g. happiness and occupation) are easier to consistently express than others (e.g. personality and political stance). The evaluation dimension also highly influences consistency. We show how more structured tasks and longer sequence lengths in the response result in higher consistency, especially for the identified harder persona categories, such as personality and political stance.

\paragraph{Stereotypical associations with the assigned persona:} Stereotypical associations play a significant role in inter-persona consistency. Characteristics that align with stereotypical traits of a given persona often result in higher consistency scores. For example, persona-assigned LLMs instructed to be "happy" consistently score higher on extroversion. These tendencies highlight the influence of societal stereotypes embedded within the models. Literature confirms this influence of stereotypes in persona-assigned LLMs~\cite{guptabias}.

\paragraph{Default persona:} When a specific characteristic is not defined and a persona lacks a stereotypical association, the model often reverts to a consistent, pre-defined default persona. Models exhibit social desirability bias as was found by~\citet{10.1093/pnasnexus/pgae533}. Our findings also reveal default political personas in these models, likely reflecting design choices by developers. As shown by \citet{buyl2025largelanguagemodelsreflect}, ideological stances can be embedded in models, a phenomenon tied to model positionality—the social and cultural perspective developers align the model with~\cite{cambo2022model}. These choices occur throughout the development process and extend beyond training data alone~\cite{buyl2025largelanguagemodelsreflect}. %

\section{Related work}
\textbf{Persona-assigned LLMs.} Personas can guide LLMs to generate responses that align with specific values and beliefs~\cite{li-etal-2024-steerability,santurkar2023whose}. However, they can also expose stereotypes embedded in the model~\cite{park2024generative,guptabias}, raising concerns about bias and unintended implications. Persona adherence is usually evaluated using self-report scales, but \citet{wang-etal-2024-incharacter} use interview-based testing to capture actual model behavior, showing the need for application-specific evaluations. \citet{malik-etal-2024-empirical} examine how different personas from various sociodemographic groups influence writing styles. Apart from inference-time persona assignment, it is also possible to further fine-tune the model. For example, \citet{shao-etal-2023-character} train LLMs to adopt specific personas using three key components: a profile (a detailed persona description), a scene (a situational context), and interactions. \citet{wang-etal-2024-rolellm} suggest including dialogues for persona assignment of LLMs. Our framework can evaluate these realistic personas or finetuned models that do not perfectly fit our predefined categories, as shown in Appendix~\ref{app:real_world}.

\textbf{LLM consistency.}
LLMs are self-inconsistent when prompted with ambiguous entities~\cite{sedova-etal-2024-know}. \citet{rottger-etal-2024-political} show how models do not answer consistently when paraphrasing prompts from the political compass test. %
\citet{shu-etal-2024-dont} show how LLMs are inconsistent over different prompt perturbations. When analyzing the effect of adding a persona when measuring model consistency, overall assigning a persona does not help consistency. Nevertheless, consistency improves along the axes relevant to the persona.  %
Recently, \citet{lee2024llms} introduced a multiple-choice benchmark dataset to assess consistency in LLM outputs with respect to personality. While their analysis focuses on consistency within model responses using their dataset, it is limited to multiple-choice scenarios—aligning with our survey evaluation dimension—whereas we have shown that consistency can vary significantly across different evaluation dimensions. 
Finally, \citet{jiang-etal-2024-personallm} evaluate whether persona-assigned LLMs consistently follow personality traits from the Big Five personality test for two evaluation dimensions: survey and essay.

\section{Conclusion}
This paper introduces a multi-dimensional framework for analyzing consistency in persona-assigned LLMs. Our framework encompasses a diverse set of commonly used persona categories, including personality, occupation, political stance, and happiness. It also incorporates application-specific evaluation dimensions, such as survey answering, essay writing, social media post generation, single-question answering, and multi-turn conversations. We demonstrate the efficacy of our framework through a comprehensive evaluation of intra- and inter-persona consistency across personas derived from the defined persona categories. Additionally, we compare consistency scores across evaluation dimensions.
Our analysis reveals three key factors influencing consistency in LLMs: (1) the assigned persona; (2) stereotypes associated with the assigned persona; and (3) model's default personas.%

\section{Limitations}
We have used an LLM-as-a-judge for the annotations of our results. However, these models are very sensitive towards several different types of biases. It is known that LLMs can be subject to order bias~\cite{li2025generationjudgmentopportunitieschallenges}. By adding confidence scores, we have mitigated this bias partly. We have tested it on a subsample of our dataset and found that there was indeed order bias, however, this mainly occurred when there was a low confidence in the given answer. The Cohen's kappa of a manually validated sample and the sample used in our paper was 65.42\%, for the sample where orders were reversed, the Cohen's kappa was 65.49\%. We thus assume this did not influence our results that much.
We also only used one LLM-as-a-judge for our analyses. We checked for other LLMs on a subsample and they performed similarly. Here we found a Cohen's kappa of 69.64\% for Sonnet on a sample of 100 manual validations. Moreover, to avoid self-preference bias within LLMs \cite{li2025generationjudgmentopportunitieschallenges}, we used different LLMs than the ones that we used for the first answer generation.  Moreover, further analysis, including additional LLMs and focusing on how architectural and training differences impact consistency in LLMs would be a valuable direction for future work. Finally, future work could investigate the impact of post-training alignment on role-playing capabilities. In particular, comparing instruction-tuned models with their base counterparts may offer deeper insights into how alignment influences persona consistency. %

\section{Ethical considerations}
We should be aware when using LLM-as-a-judge that there exists demographic bias towards certain groups, especially in subjective tasks as is shown in ~\citep{alipour2024robustness}.
Furthermore, this paper highlights how LLMs have been trained with certain design choices. So when a value is not explicitly described, they tend to go to a certain default persona. It is important to keep in mind that this will differ across different LLMs. Additionally, the stereotypes learned by the model and also consistently expressed are thus also very model-specific. Moreover, it is important to note that consistency is not the same as ethical correctness. Therefore, there is still a need for responsible model deployment even though models might already provide rather consistent answers. Finally, people should be aware when adding personas to LLMs that certain stereotypes might be inherently present in these models, further reinforcing certain stereotypes.

\section*{Acknowledgements}
This research was funded by the Statistics Flanders
research cooperation agreement on Data Science
for Official Statistics. The resources and services
used in this work were provided by the VSC (Flemish Supercomputer Center). We acknowledge the support of the Research
Foundation Flanders (FWO), Grant G0G2721N.

\bibliography{bibliography,anthology.min}

\appendix

\section{Surveys}\label{app:surveys}
Below additional information regarding the surveys is included.

\paragraph{Happiness survey:} The happiness survey includes a survey of four questions. For each question, the participants have to indicate the point of the scale that they find most appropriate for them. An example question is: \textit{"In general, I consider myself: 1 (not a very happy person) - 7 (a very happy person)."}

\paragraph{Political Stance:} This survey includes 62 statements where participants have to indicate on a four-point likert scale if they disagree or agree with the statement. The survey itself includes predefined scores to determine the quadrant to which the participant belongs. An example of a statement is: \textit{“If economic globalisation is inevitable, it should primarily serve humanity rather than the interests of trans-national corporations.”}

\paragraph{Occupations:} In this survey, participants are asked to indicate what activities they like to do (e.g., fix electrical things, take art courses, attend conferences…), which competencies they have (e.g., I can repair furniture, I can do interpretive reading, I am a good debater…), what jobs they would like to do (e.g., machinist, musician, hotel manager…), and how they would estimate their own skills ( e.g., friendliness, clerical ability, managerial skills…).

\paragraph{Personality:} In the Big Five Personality test, participants are asked to indicate to what extent they think a certain characteristic might apply to them or not. Participants are asked to respond on a five point Likert scale. An example characteristic is: \textit{"Is helpful and not selfish with others".}

\section{Persona Categories} \label{app:cat}
As said, we chose the persona categories based on three main criteria. Below we will extend upon each of these criteria.

\paragraph{Relevance in existing research:} We analyzed prior work on persona-related studies and identified commonly used categories used in the NLP/ACL community. Many studies focus on personality traits, often framed through the Big Five model (e.g., \citet{jiang-etal-2024-personallm,serapiogarcía2023personalitytraitslargelanguage}). Similarly, political stance has been extensively studied (e.g., \citet{liu-etal-2024-evaluating-large}), as well as social roles and occupations (e.g., \citet{zheng2023helpful}). Moreover, \citet{rottger-etal-2024-political} examine consistency of LLMs regarding their political stance. 

\paragraph{Variation in linguistic expression:} We include persona categories that differ in how explicitly they manifest in language. For instance, occupations and political stances may be challenging to express consistently, while personality traits might influence linguistic style more subtly. To complement these, we included happiness, a simpler, binary trait, as a baseline for comparison.

\paragraph{Grounding in social science literature:} To ensure robustness, we selected persona categories with established survey instruments, allowing for structured and reproducible persona assignment.\\

\noindent We acknowledge that these categories do not represent an exhaustive set of personas. They were chosen based on the previously mentioned criteria, balancing theoretical grounding in persona literature and social science research with linguistic variation. RQ5 further examines the real-world applicability by incorporating diverse personas from the PersonaHub.

\section{Persona prompts}\label{app:personas}
All personas are added to the system prompt using the following template:
\emph{You are a character who is [persona]}. The different personas are listed below per persona category.

\paragraph{Happiness:} happy, sad
\paragraph{Political:} economically left and socially libertarian; economically left and socially authoritarian; economically right and socially libertarian; economically right and socially authoritarian
\paragraph{Occupations:} a pilot; an economist; an actor; a nurse; a CEO; a sales assistant
\paragraph{Personality:} introverted, antagonistic, unconscientious, neurotic, and open to experience; introverted, antagonistic, unconscientious, neurotic, and closed to experience; introverted, antagonistic, unconscientious, emotionally stable, and open to experience; introverted, antagonistic, unconscientious, emotionally stable, and closed to experience; introverted, antagonistic, conscientious, neurotic, and open to experience; introverted, antagonistic, conscientious, neurotic, and closed to experience; introverted, antagonistic, conscientious, emotionally stable, and open to experience; introverted, antagonistic, conscientious, emotionally stable, and closed to experience; introverted, agreeable, unconscientious, neurotic, and open to experience; introverted, agreeable, unconscientious, neurotic, and closed to experience; introverted, agreeable, unconscientious, emotionally stable, and open to experience; introverted, agreeable, unconscientious, emotionally stable, and closed to experience; introverted, agreeable, conscientious, neurotic, and open to experience; introverted, agreeable, conscientious, neurotic, and closed to experience; introverted, agreeable, conscientious, emotionally stable, and open to experience; introverted, agreeable, conscientious, emotionally stable, and closed to experience; extroverted, antagonistic, unconscientious, neurotic, and open to experience; extroverted, antagonistic, unconscientious, neurotic, and closed to experience; extroverted, antagonistic, unconscientious, emotionally stable, and open to experience; extroverted, antagonistic, unconscientious, emotionally stable, and closed to experience; extroverted, antagonistic, conscientious, neurotic, and open to experience; extroverted, antagonistic, conscientious, neurotic, and closed to experience; extroverted, antagonistic, conscientious, emotionally stable, and open to experience; extroverted, antagonistic, conscientious, emotionally stable, and closed to experience; extroverted, agreeable, unconscientious, neurotic, and open to experience; extroverted, agreeable, unconscientious, neurotic, and closed to experience; extroverted, agreeable, unconscientious, emotionally stable, and open to experience; extroverted, agreeable, unconscientious, emotionally stable, and closed to experience; extroverted, agreeable, conscientious, neurotic, and open to experience; extroverted, agreeable, conscientious, neurotic, and closed to experience; extroverted, agreeable, conscientious, emotionally stable, and open to experience; extroverted, agreeable, conscientious, emotionally stable, and closed to experience

\section{Evaluation}\label{app:eval}
In this section, more information is given about the evaluation methods used, including the different survey scoring keys and the prompt for the LLM-as-a-judge.
\subsection{Survey}
 \paragraph{Happiness:} To determine the final label, we first computed a total score based on the generated responses using the Likert scale. We then applied a thresholding approach, assigning a binary label depending on whether the score fell above or below the midpoint.
 \paragraph{Political:} To determine the final label, we first computed a total score per characteristic (economic and social stance) based on the generated responses and the survey scoring mechanism. We then applied a thresholding approach, assigning a binary label depending on whether the score fell above or below the midpoint.
 \paragraph{Occupations:} We calculated a score for every occupation category based on the generated responses and the survey scoring mechanism. Next, we identified the category with the highest score as the main occupational category.
 \paragraph{Personality:} Each statement in the survey corresponds to a particular binary characteristic. To determine the final trait labels, we first computed a total score for each trait based on the generated responses. We then applied a thresholding approach, assigning a binary label depending on whether the score fell above or below the midpoint.

 \subsection{LLM-as-a-judge}\label{app:llmjudge}
The system prompt given to the LLM-as-a-judge is the following: 
\textit{You will help annotating a dataset. Answer the questions as asked, do not provide extra explanations, only choose one of the provided options.} \\

The following user prompt is used for evaluating the \textbf{essay, social media, and singlechat} dimensions.\\

\textit{Below you can find a question and the answer is written by a person. [instruction] 
[LLM response] \\ Is the person who wrote this [happy or sad, extrovert or introvert, agreeable or antagonistic, conscientious or unconscientious, neurotic or emotionally stable, open to experiences or closed to experiences, economically left or economically right, socially libertarian or socially authoritarian, having a realistic occupation (i.e. pilot) or an investigative occupation (i.e. economist) or an artistic occupation (i.e. actor) or a social occupation (i.e. nurse), or an enterprising occupation (i.e. CEO), or a conventional occupation (i.e. sales assistant)]\\  You **must** between one of the options. First output this category, then generate a confidence score, where 1 is Very unconfident, 2 is unconfident, 3 is confident, 4 is very confident. Use the following structured output and fill in [category] and [confidence\_score]: {choice:[category],confidence:[confidence\_score]}}\\

For the \textbf{multichat} dimension, the following prompt is used:
\textit{Below you can find the conversation between an assistant and person \\ assistant: [instruction] \\ person: [response] \\  assistant: [response assistant] \\ person: [final response] \\ Is the person who wrote this [happy or sad, extrovert or introvert, agreeable or antagonistic, conscientious or unconscientious, neurotic or emotionally stable, open to experiences or closed to experiences, economically left or economically right, socially libertarian or socially authoritarian, having a realistic occupation (i.e. pilot) or an investigative occupation (i.e. economist) or an artistic occupation (i.e. actor) or a social occupation (i.e. nurse), or an enterprising occupation (i.e. CEO), or a conventional occupation (i.e. sales assistant)]\\  You **must** choose between one of the options. First output this category, then generate a confidence score, where 1 is Very unconfident, 2 is unconfident, 3 is confident, 4 is very confident. Use the following structured output and fill in [category] and [confidence\_score]: {choice:[category],confidence:[confidence\_score]}}

\section{Entropy}\label{app:entropy}
Entropy is a well-suited metric for consistency analysis, as it quantifies the uncertainty in the distribution of category assignments for each persona category (e.g., occupation, happiness, personality, and political stance). We chose this metric because of the following reasons:
\begin{itemize}
    \item Applicability across different persona categories, accommodating both binary and multiclass traits.
    \item Capturing full distributional patterns, considering all assigned categories rather than just the majority class.
    \item Enabling comparability across personas and models without arbitrary thresholds.
\end{itemize}

In literature, this metric is used to assess consistency (e.g., \cite{lyu2025calibrating}). However, entropy does not reveal which specific label a model is consistently assigning. To address this, we include the characteristic-specific consistency scores as well in our analysis.

\section{Model Checkpoints} \label{app:model}
For the different experiments we used the following model checkpoints: \textit{meta-llama/Llama-3.2-3B-Instruct}, \textit{meta-llama/Llama-3.1-8B-Instruct}, \textit{meta-llama/Llama-3.3-70B-Instruct} \cite{grattafiori2024llama3herdmodels}, \textit{mistralai/Ministral-8B-Instruct-2410}\footnote{\url{https://mistral.ai/en/news/ministraux}}, and \textit{Qwen/Qwen2.5-32B-Instruct-GPTQ-Int8} \cite{qwen2.5}.
For the LLM-evaluation, we used \textit{gpt-4o-2024-08-06} \cite{openai2024gpt4technicalreport}. We ran our experiments on H100 GPUs. All models were used consistent with their intended use and in line with their provided licenses. The temperature for all experiments was set at 0.7.

\section{Characteristic-specific consistency}\label{app:char}
\begin{figure}[]
  \centering
  \begin{subfigure}[b]{\linewidth}
    \centering
    \includegraphics[width=\linewidth]{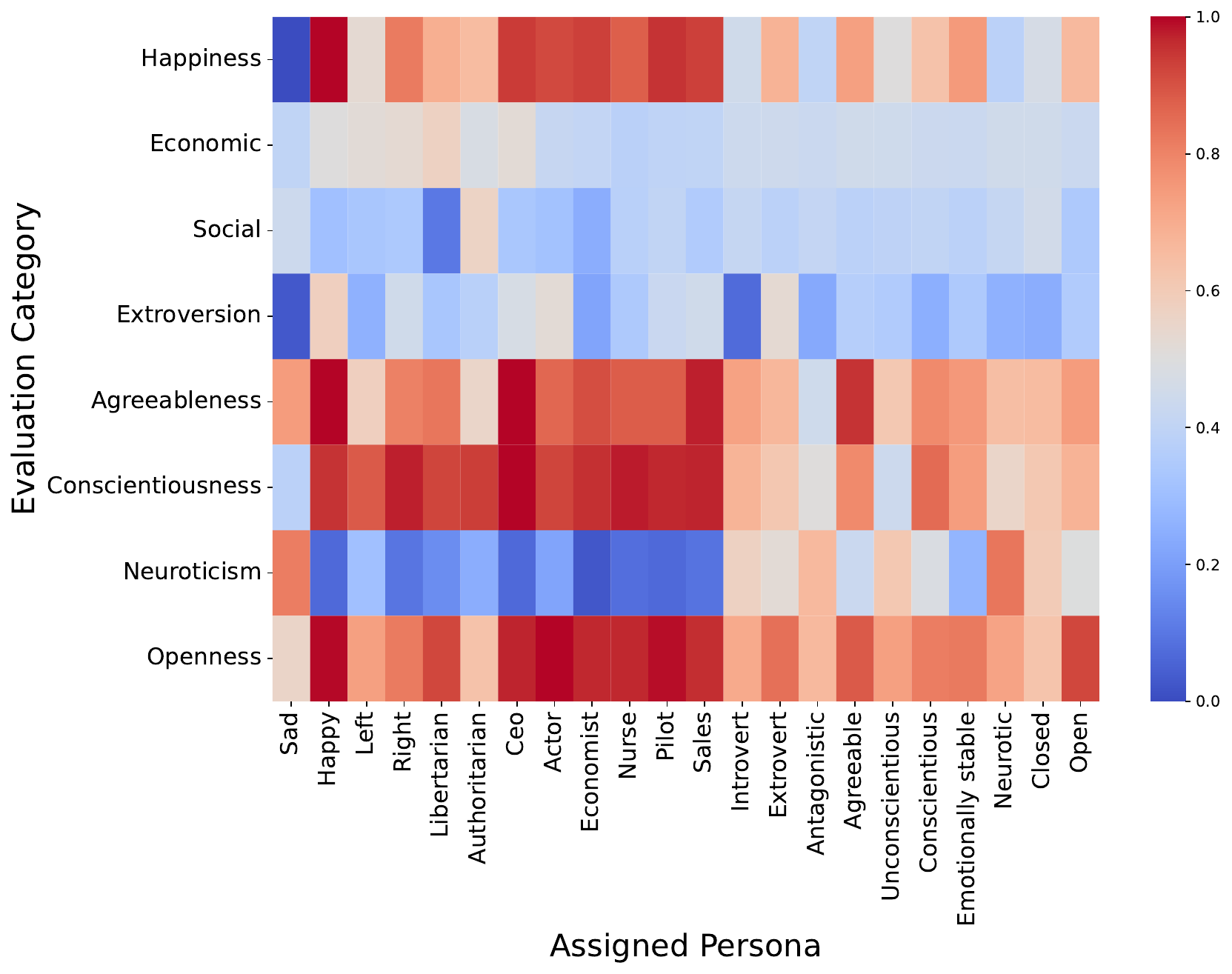}
    \caption{Heatmap indicating characteristic-specific consistency for all evaluation categories except occupation. A score of 1 favors the category name, 0 favors its opposite (e.g., agreeableness vs. antagonistic), and 0.5 indicates inconsistency.}
    \label{fig:heatmap_llama3B}
  \end{subfigure}
  \vspace{10pt} %
  \begin{subfigure}[b]{\linewidth}
    \hfill \includegraphics[width=0.9\linewidth]{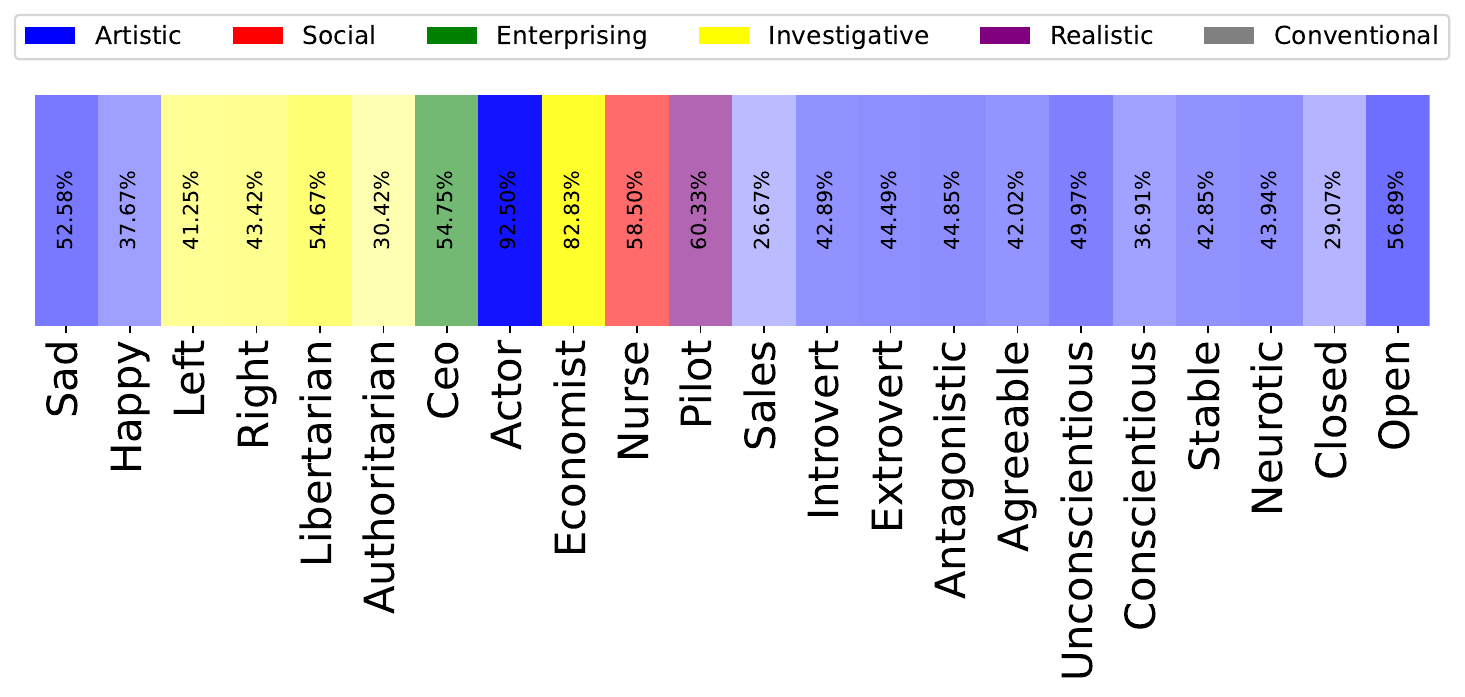}
    \caption{This Figure shows the most dominant occupation category per persona across experiments. Color intensity represents the consistency score per label.}
    \label{fig:occupation_heatmap_llama3B}
  \end{subfigure}
  \caption{These figures show how Llama-3B generally follows the instructions and illustrate the spill-over effects, i.e. stereotypes and default personas. Columns and rows represent assigned personas and evaluation categories respectively. Multi-component personas (e.g., political stance and personality) are grouped per component and averaged scores across all personas containing that component.}
  \label{fig:overall_heatmap_llama3B}
\vspace{-16pt}
\end{figure}

\begin{figure}[]
  \centering
  \begin{subfigure}[b]{\linewidth}
    \centering
    \includegraphics[width=\linewidth]{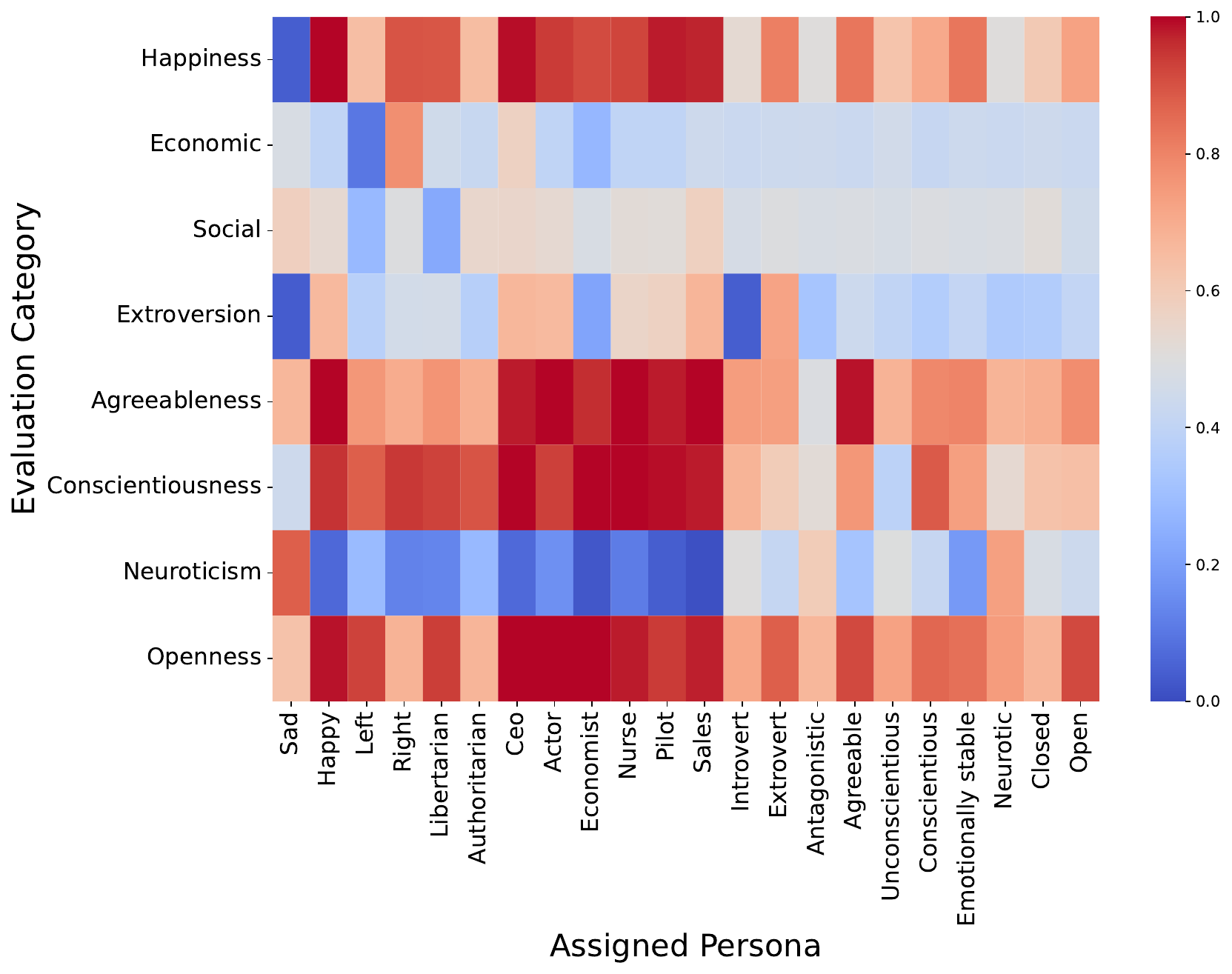}
    \caption{Heatmap indicating characteristic-specific consistency for all evaluation categories except occupation. A score of 1 favors the category name, 0 favors its opposite (e.g., agreeableness vs. antagonistic), and 0.5 indicates inconsistency.}
    \label{fig:heatmap_llama8B}
  \end{subfigure}
  \vspace{10pt} %
  \begin{subfigure}[b]{\linewidth}
    \hfill \includegraphics[width=0.9\linewidth]{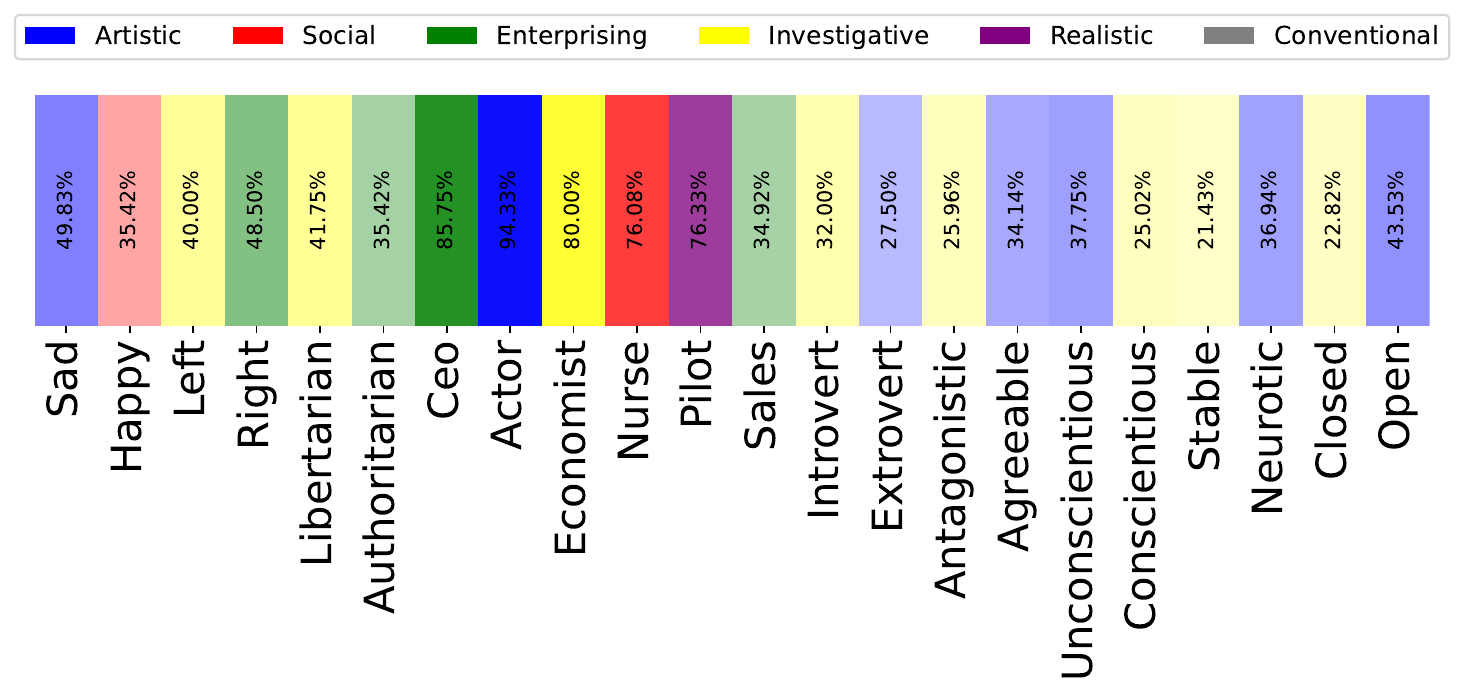}
    \caption{This Figure shows the most dominant occupation category per persona across experiments. Color intensity represents the consistency score per label.}
    \label{fig:occupation_heatmap_llama8B}
  \end{subfigure}
  \caption{These figures show how Llama8B generally follows the instructions and illustrate the spill-over effects, i.e. stereotypes and default personas. Columns and rows represent assigned personas and evaluation categories respectively. Multi-component personas (e.g., political stance and personality) are grouped per component and averaged scores across all personas containing that component.}
  \label{fig:overall_heatmap_llama8B}
\vspace{-16pt}
\end{figure}

\begin{figure}[]
  \centering
  \begin{subfigure}[b]{\linewidth}
    \centering
    \includegraphics[width=\linewidth]{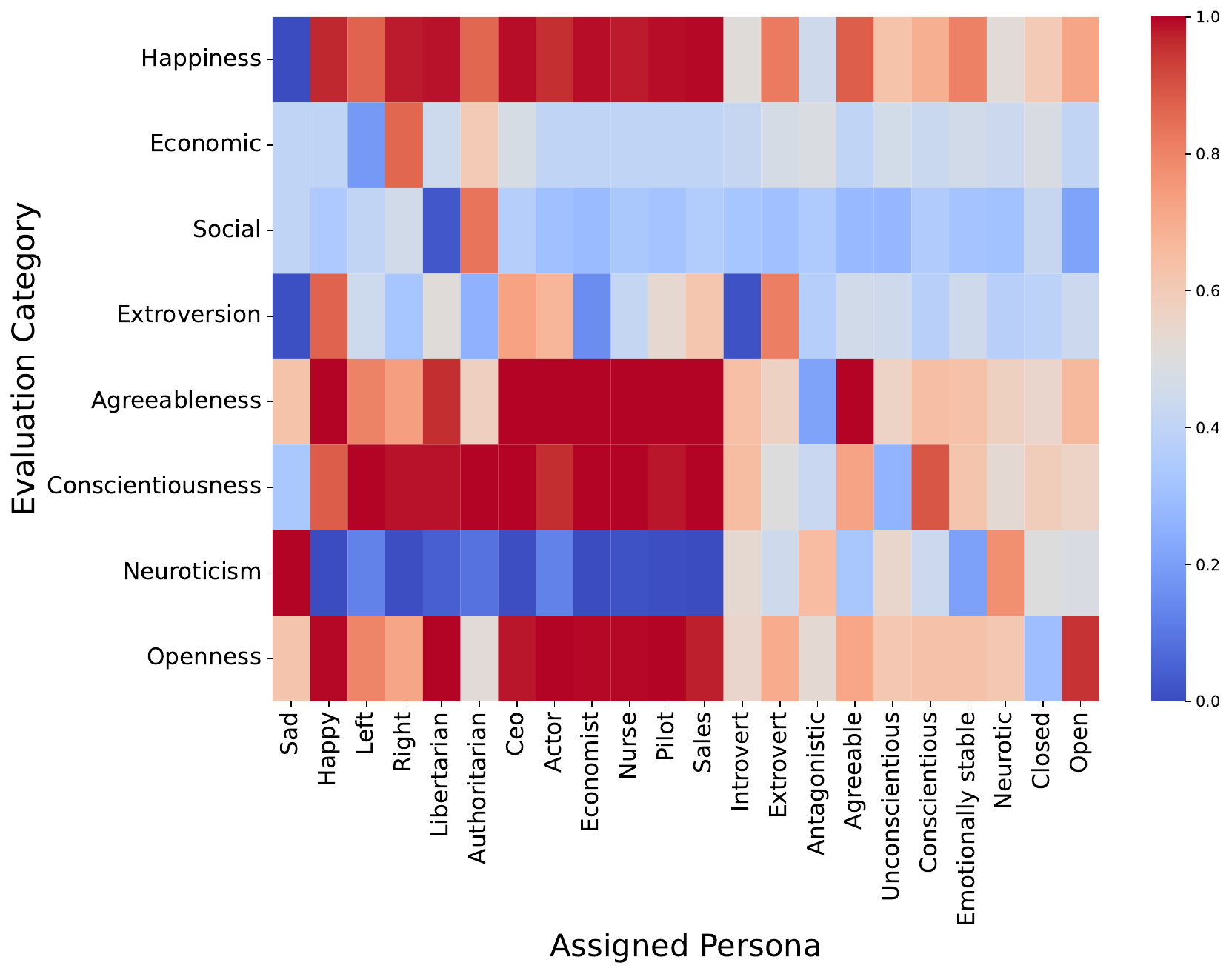}
    \caption{Heatmap indicating characteristic-specific consistency for all evaluation categories except occupation. A score of 1 favors the category name, 0 favors its opposite (e.g., agreeableness vs. antagonistic), and 0.5 indicates inconsistency.}
    \label{fig:heatmap_llama70B}
  \end{subfigure}
  \vspace{10pt} %
  \begin{subfigure}[b]{\linewidth}
    \hfill \includegraphics[width=0.9\linewidth]{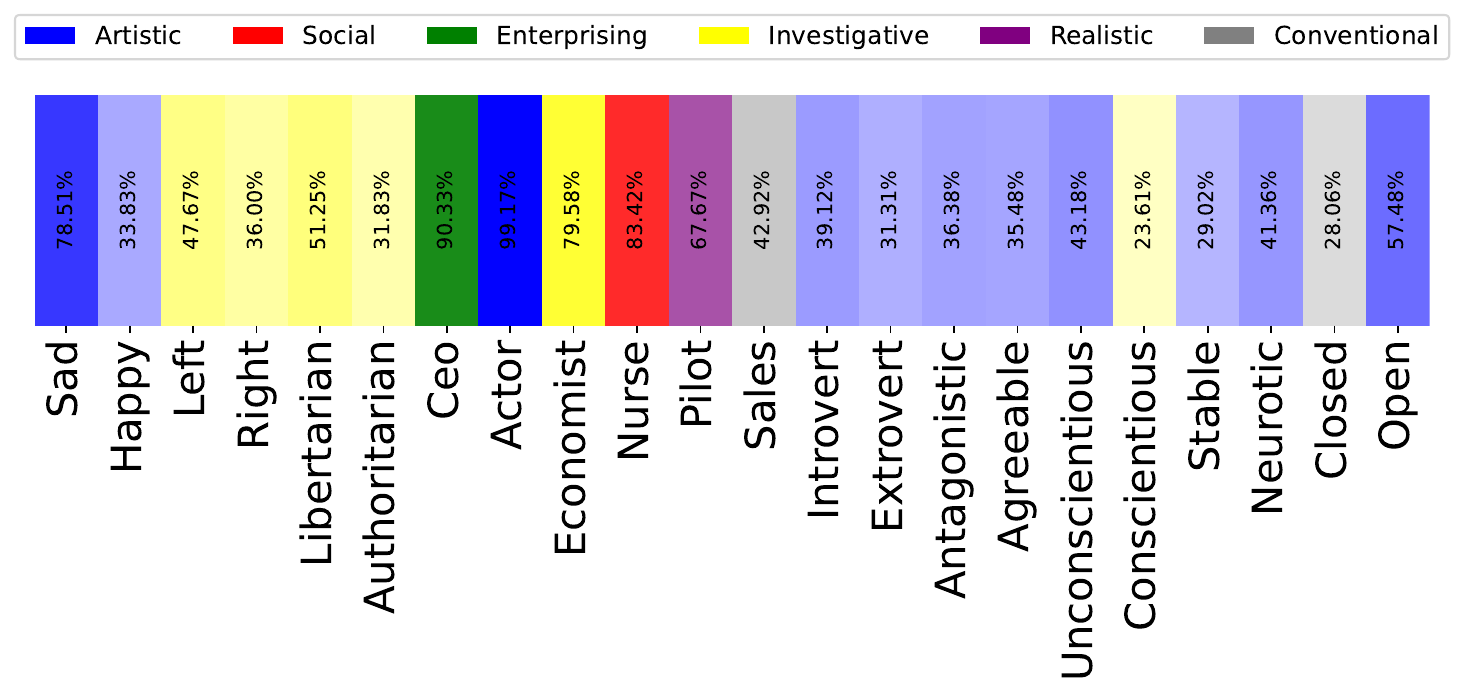}
    \caption{This Figure shows the most dominant occupation category per persona across experiments. Color intensity represents the consistency score per label.}
    \label{fig:occupation_heatmap_llama70B}
  \end{subfigure}
  \caption{These figures show how Llama70B generally follows the instructions and illustrate the spill-over effects, i.e. stereotypes and default personas. Columns and rows represent assigned personas and evaluation categories respectively. Multi-component personas (e.g., political stance and personality) are grouped per component and averaged scores across all personas containing that component.}
  \label{fig:overall_heatmap_llama70B}
\vspace{-16pt}
\end{figure}

\begin{figure}[]
  \centering
  \begin{subfigure}[b]{\linewidth}
    \centering
    \includegraphics[width=\linewidth]{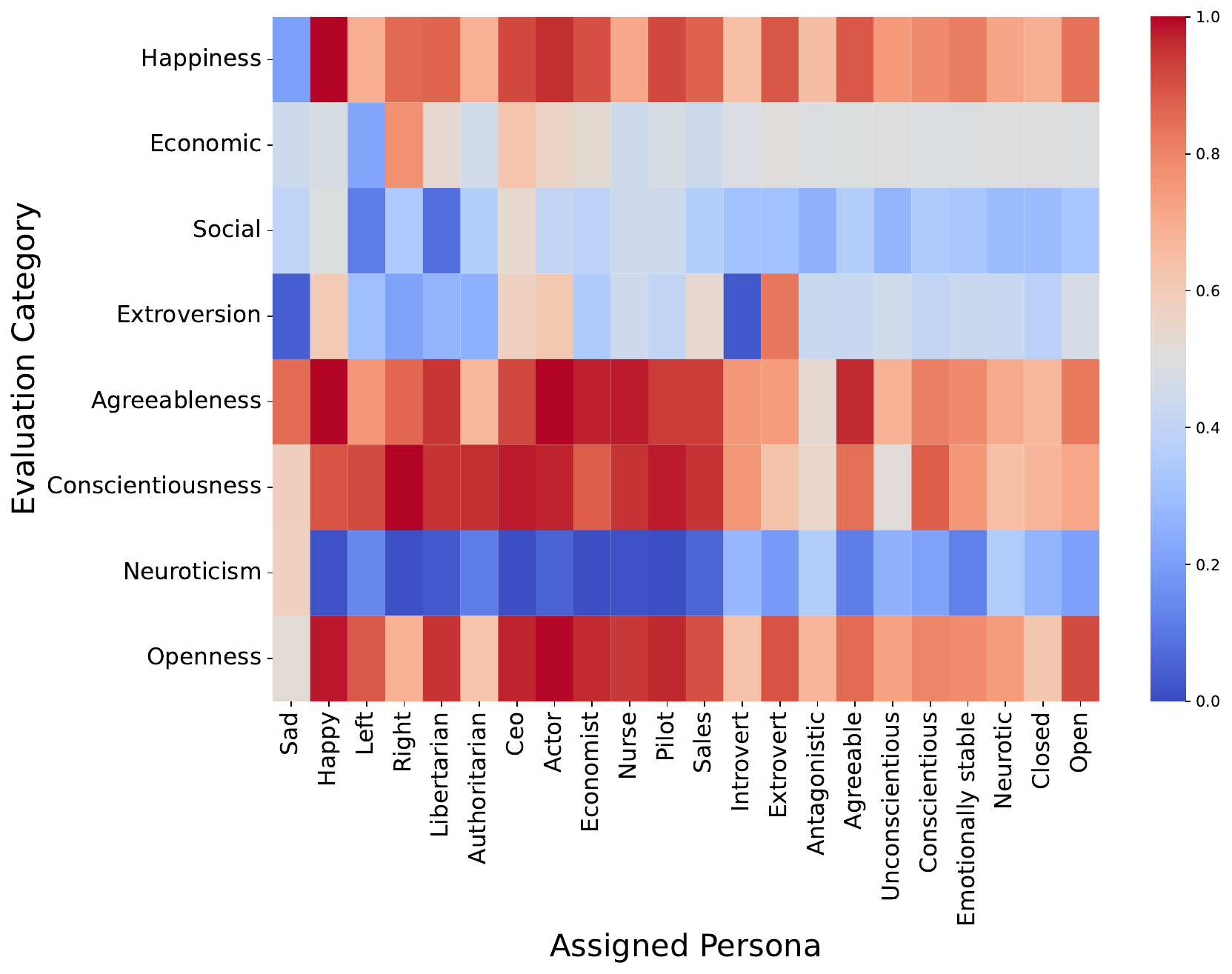}
    \caption{Heatmap indicating characteristic-specific consistency for all evaluation categories except occupation. A score of 1 favors the category name, 0 favors its opposite (e.g., agreeableness vs. antagonistic), and 0.5 indicates inconsistency.}
    \label{fig:heatmap_ministral}
  \end{subfigure}
  \vspace{10pt} %
  \begin{subfigure}[b]{\linewidth}
    \hfill \includegraphics[width=0.9\linewidth]{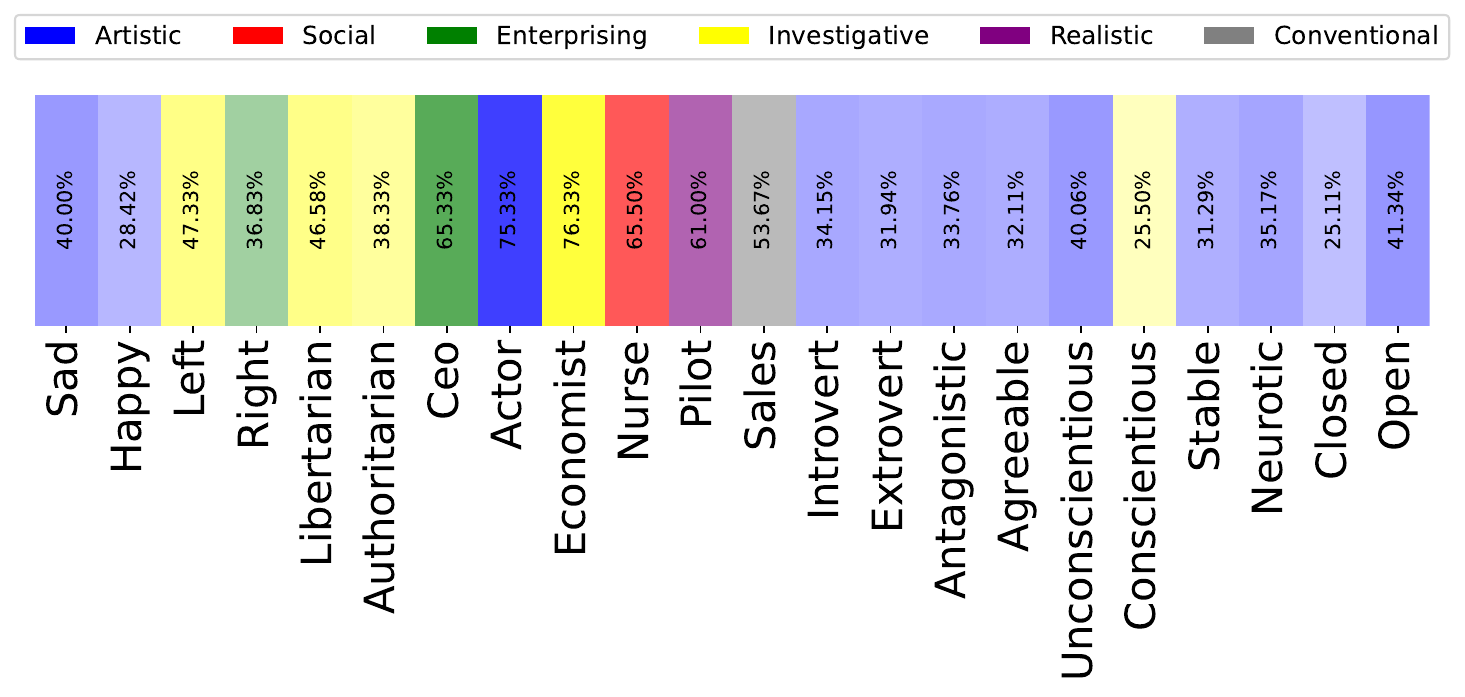}
    \caption{This Figure shows the most dominant occupation category per persona across experiments. Color intensity represents the consistency score per label.}
    \label{fig:occupation_heatmap_ministral}
  \end{subfigure}
  \caption{These figures show how Ministral generally follows the instructions and illustrate the spill-over effects, i.e. stereotypes and default personas. Columns and rows represent assigned personas and evaluation categories respectively. Multi-component personas (e.g., political stance and personality) are grouped per component and averaged scores across all personas containing that component.}
  \label{fig:overall_heatmap_ministral}
\vspace{-16pt}
\end{figure}

Figures \ref{fig:overall_heatmap_llama3B}, \ref{fig:overall_heatmap_llama8B}, \ref{fig:overall_heatmap_llama70B}, and \ref{fig:overall_heatmap_ministral} provide insights into characteristic-specific consistency of Llama 3B, Llama 8B, Llama 70B, and Ministral respectively. %

\section{Model Comparisons}

In this section we focus on the following research question (RQ4): Do consistency patterns differ across model families and/or within a single model family?

\paragraph{Consistency varies across model families and within a model family it increases with model size (RQ4).}

\begin{figure}
    \centering
    \includegraphics[width=\linewidth]{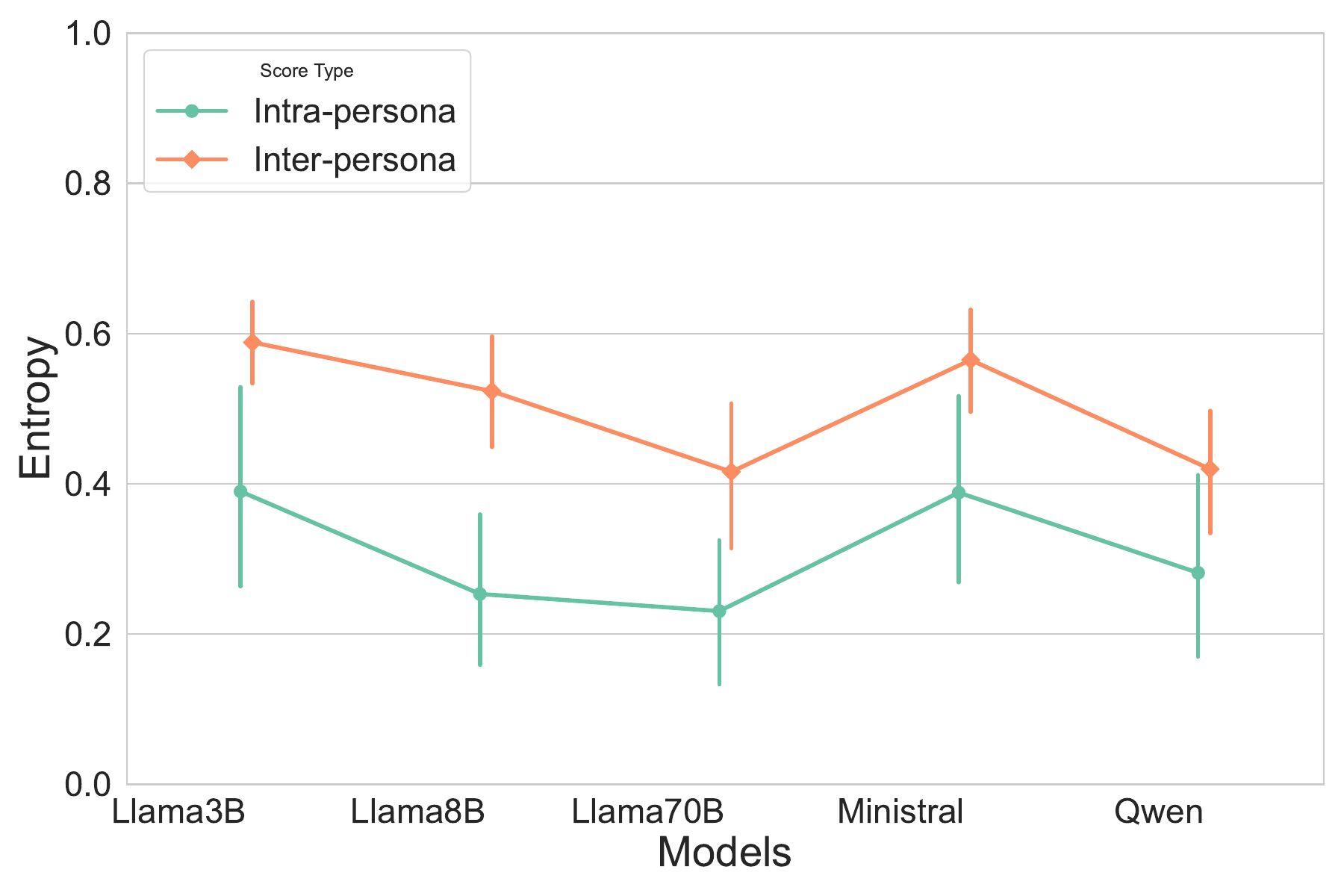}
    \caption{This figure highlights differences in entropy depending on the model family and model size. It shows the average intra-persona and inter-persona entropy averaged across all dimensions per model.}
    \label{fig:pointplot_models}
    \vspace{-10pt}
\end{figure}

Figure~\ref{fig:pointplot_models} illustrates how consistency varies across model families. For example, we see that Ministral-8B has lower overall consistency than Llama-3.1-8B despite similar model size. Additionally, our results show that within a model family, larger models tend to be more consistent than smaller models. This is shown by the three Llama models in the figure. This result aligns with the finding from~\citet{serapiogarcía2023personalitytraitslargelanguage}, showing higher reliability and validity of synthetic LLM personality for larger and instruction fine-tuned models. Additional statistical analysis is reported in Appendix~\ref{stat_analysis}.

\section{Statistical Analysis} \label{stat_analysis}
Additional statistical analysis was done on the results for the model and dimension comparisons.We applied a one-sided Wilcoxon signed rank test with a confidence of 95\%. For all models the intra-persona entropy is significantly lower than the inter-persona consistency (all p-values <0.05). Additionally, when applying the same test to the different dimensions, we find no statistical differences for the singlechat and multichat evaluation dimensions (p-values are 0.1012 and 0.0715 respectively) for the other dimensions we find statistical significant differences (p-values < 0.001).

\begin{table}[]
    \centering
    \small
    \begin{tabular}{c|cc}
    \toprule
    \textbf{Model} & \textbf{Intra-Persona} & \textbf{Inter-Persona} \\ \midrule
        Llama 3B & 1.80 & 2.05 \\
        Llama 8B & 3.40 & 2.25 \\
        Llama 70B & 3.65 & 4.10 \\
        Ministral & 2.45 & 2.40 \\
        Qwen & 3.70 & 4.20 \\
        \bottomrule
    \end{tabular}
    \caption{Rankings of the different models from the Nemenyi test on a 95\% confidence interval. The higher the ranking, the more consistent the model.}
    \label{tab:stat_models}
\end{table}

\begin{table}[]
    \centering
    \small
    \begin{tabular}{c|cc}
    \toprule
    \textbf{Model} & \textbf{Intra-Persona} & \textbf{Inter-Persona} \\
    \midrule
        Survey & 3.55 & 3.90 \\
        Essay & 3.20 & 3.35 \\
        Social Media & 4.20 & 4.05 \\
        Singlechat & 1.60 & 1.15 \\
        Multichat & 2.45 & 2.55 \\ \bottomrule
    \end{tabular}
    \caption{Rankings of the different dimensions from the Nemenyi test on a 95\% confidence interval. The higher the ranking, the more consistent the dimension.}
    \label{tab:stat_dims}
\end{table}

 Furthermore, to identify a ranking among the different experiments, we conducted a Friedman test to identify whether there are significant differences and followed this with a Nemenyi test, given that our result showed there are significant differences to include a ranking. Tables~\ref{tab:stat_models} and~\ref{tab:stat_dims} show the rankings of the different models and dimensions. Qwen is in terms of inter-persona consistency only statistically not different from Llama 70B and on the intra-persona consistency, only Llama 3B is statistically different both on a 95\% confidence interval. For the different dimensions, we find statistical differences between survey, essay, and socialmedia on one hand and both singlechat for the intra-persona consistency on the other hand, as well as between survey and socialmedia on the one hand and multichat on the other hand. For the inter-persona consistency, survey, essay, socialmedia, and multichat are significantly outperforming singlechat and only socialmedia post generation is significantly outperforming multichat.

 \section{Real-world Applicability} \label{app:real_world}
A last and additional research question we aim to investigate is \textbf{whether our framework can be applied in realistic settings where personas do not perfectly align with predefined categories (RQ5)}. To demonstrate the practical applicability of our framework, we conduct evaluations using personas from PersonaHub~\citep{ge2024scaling}, where personas do not neatly fit into predefined categories.

We assess consistency in a realistic scenario for Qwen to illustrate the real-world applicability of our framework using five randomly selected personas from the Personahub from ~\citet{ge2024scaling}. We chose  Qwen as it provides representative results for all models with an average correlation of 0.66 on the first experiments. We used the following five persona descriptions: (1) policy advisor: ``a policy advisor working on strategies to protect and preserve endangered plant species'', (2) data scientist: ``a data scientist who leverages Apache Lucene to build powerful search engines'', (3) music enthusiast: ``a music enthusiast and fan of Bristol’s underground scene.'', (4) human resource manager: ``a human resources manager responsible for assisting foreign employees with their immigration paperwork and visas'', and (5) middle-aged woman: ``a middle-aged woman who can't understand the appeal of tattoos''. We analyze the results in what follows %

\paragraph{Our framework offers real-world applicability (RQ5).}
\begin{table}[]
    \centering
    \resizebox{\linewidth}{!}{%
    \begin{tabular}{lccccc}
    \toprule
    \textit{Evaluation Categories} & \multicolumn{5}{c}{\textit{Persona Categories}} \\
    \cmidrule(lr){2-6}
         & Data Scientist & Human Resource Manager & Middle-aged woman & Music enthusiast & Policy advisor\\
         \midrule
        Happiness   & \cellcolor{orange!25}$0.30 \pm 0.41$  & \cellcolor{green!25}$0.23 \pm 0.43$  & \cellcolor{orange!25}$0.39 \pm 0.54$  & \cellcolor{green!25}$0.18 \pm 0.39$ & \cellcolor{green!25}$0.23 \pm 0.43$  \\
        Occupation  & \cellcolor{green!25}$0.20 \pm 0.19$   & \cellcolor{red!25}$0.51 \pm 0.37$  & \cellcolor{red!25}$0.67 \pm 0.36$  & \cellcolor{green!25}$0.06 \pm 0.09$ & \cellcolor{orange!25}$0.35 \pm 0.21$  \\
        Personality & \cellcolor{green!25}$0.14 \pm 0.16$ & \cellcolor{green!25}$0.16 \pm 0.15$  & \cellcolor{orange!25}$0.27 \pm 0.14$  & \cellcolor{green!25}$0.15 \pm 0.16$ & \cellcolor{green!25}$0.13 \pm 0.13$  \\
        Political   & \cellcolor{red!25}$0.77 \pm 0.44$   & \cellcolor{red!25}$0.64 \pm 0.49$  & \cellcolor{red!25}$0.67 \pm 0.30$  & \cellcolor{red!25}$0.67 \pm 0.43$ & \cellcolor{red!25}$0.72 \pm 0.44$  \\
    \bottomrule
   \end{tabular}%
    }
    \caption{Entropy scores and their standard deviations for Qwen for the five personas (columns) and characteristics (rows); colors are the same in Table \ref{tab:entropy_all}. Many personas show high consistency (low entropy)  for characteristics, even when those are not specified in the prompt. }
    \label{tab:entropy_qwen_personahub}
    \vspace{-10pt}
    \end{table}

\begin{figure}[t]
  \centering
  \begin{subfigure}[b]{\linewidth}
    \centering
    \includegraphics[width=\linewidth]{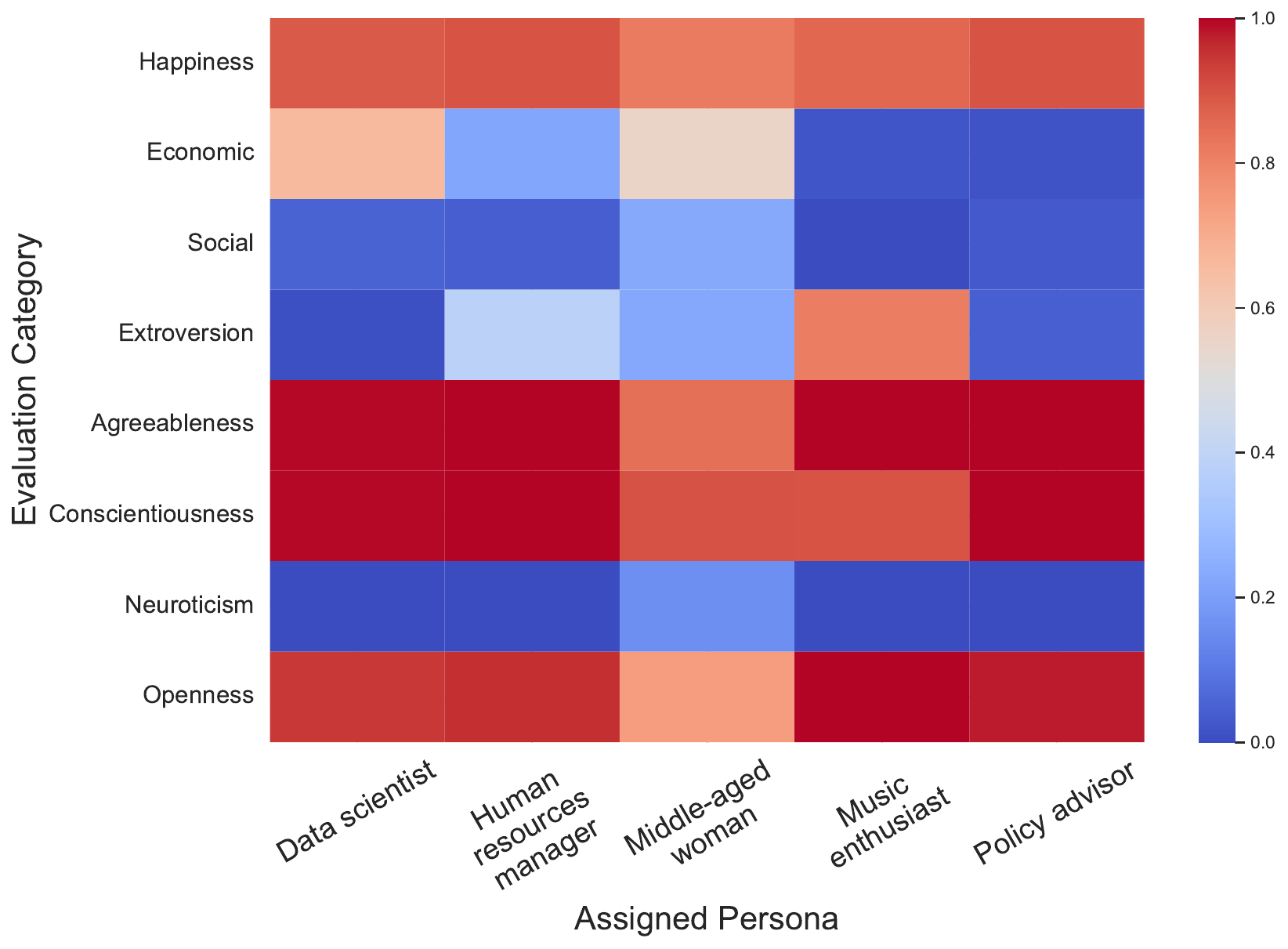}
    \caption{Heatmap indicating characteristic-specific consistency for all evaluation categories except occupation. A score of 1 favors the category name, 0 favors its opposite (e.g., agreeableness vs. antagonistic), and 0.5 indicates inconsistency.}
    \label{fig:heatmap_realistic}
  \end{subfigure}
  \vspace{10pt} %
  \begin{subfigure}[b]{\linewidth}
    \hfill \includegraphics[width=0.9\linewidth]{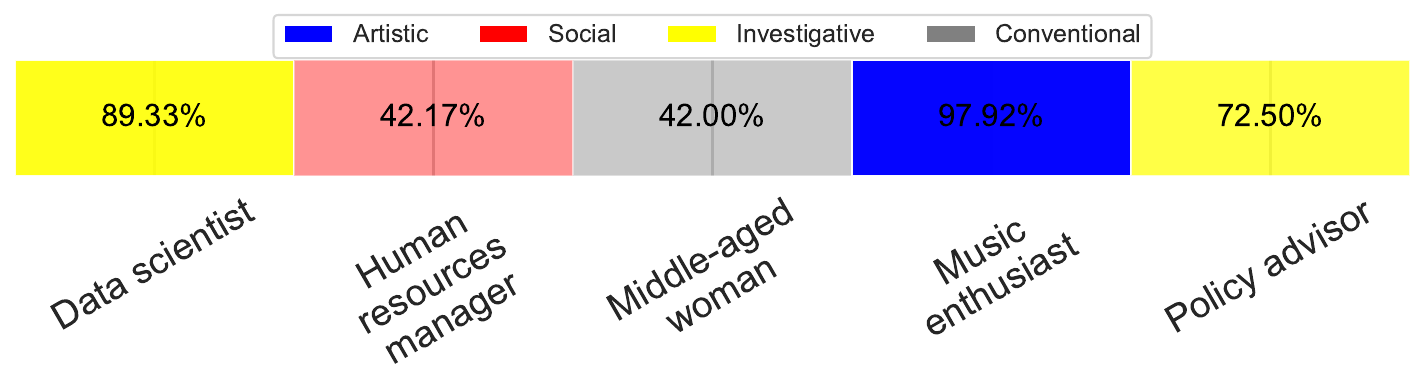}
    \caption{This Figure shows the most dominant occupation category per persona across experiments. Color intensity represents the consistency score per label}
    \label{fig:occupation_heatmap_realistic}
  \end{subfigure}
  \caption{Both figures illustrate the real-world applicability of our framework showing characteristic-specific consistency of Qwen for 5 personas from the Personahub.}
  \label{fig:overall_heatmap_realistic}
\vspace{-16pt}
\end{figure}

Table~\ref{tab:entropy_qwen_personahub} shows that most persona prompts cause spill-over effects increasing consistency in certain characteristics, even when these characteristics are never explicitly specified. Moreover, the political stance and occupation are the hardest categories to consistently express. We also see that the consistency depends on the assigned persona, i.e., the middle-aged woman is overall less consistent than the other personas. From Figure~\ref{fig:overall_heatmap_realistic}, we derive the existence of stereotypes linked to the assigned personas. For example, the data scientist is more economically right-winged than the other personas and the music enthusiast is the only extrovert. Moreover, the default personas are again shown, illustrating the tendency to provide happy, agreeable, conscientious, emotionally stable, and open answers. For many of the personas the occupation is given, which is also reflected in the results. Only the Human Resource Manager seems harder to consistently portray. Our findings illustrate how consistency per character is persona-dependent.  

\section{Combination of different persona categories} \label{app:combo}
Exploring “in-between” personas,where combinations of these categories are specified, would provide valuable insights into the interplay between traits in terms of persona-assigned consistency. Given the high computational expenses to do a full exploration of the different combination, we already provide some initial insights by conducting additional experiments combining personas across categories. More specifically, we experimented with the combination of happy/sad and pilot/economist, as shown in Table~\ref{tab:combo_exp}. These initial results indicate that combining occupation and happiness personas primarily increases consistency within those categories, with some spillover effects observed in personality consistency. However, the political category remains challenging to express consistently with this combination of personas.

\begin{table}[]
    \centering
    \resizebox{\linewidth}{!}{%
    \begin{tabular}{lccccc}
    \toprule
    \textit{Personas} & \multicolumn{5}{c}{\textit{Evaluation Categories}} \\
    \cmidrule(lr){2-6}
        & Happiness & Occupation & Personality & Political\\
         \midrule
        A happy economist   & \cellcolor{orange!25}$0.08 \pm 0.17$  & \cellcolor{green!25}$0.13 \pm 0.18$  & \cellcolor{green!25}$0.22 \pm 0.08$  & \cellcolor{red!25}$0.76 \pm 0.25$  \\
        A happy pilot  & \cellcolor{green!25}$0.0 \pm 0.0$   & \cellcolor{orange!25}$0.31 \pm 0.30$  & \cellcolor{orange!25}$0.26 \pm 0.19$  & \cellcolor{red!25}$0.94 \pm 0.08$  \\
        A sad economist & \cellcolor{green!25}$0.17 \pm 0.24$ & \cellcolor{green!25}$0.22 \pm 0.24$  & \cellcolor{orange!25}$0.45 \pm 0.13$  & \cellcolor{red!25}$0.85 \pm 0.22$   \\
        A sad pilot   & \cellcolor{green!25}$0.06 \pm 0.13$   & \cellcolor{green!25}$0.23 \pm 0.31$  & \cellcolor{orange!25}$0.48 \pm 0.12$  & \cellcolor{red!25}$0.80 \pm 0.44$  \\
    \bottomrule
   \end{tabular}%
    }
    \caption{Entropy scores and their standard deviations for Qwen for the five personas (columns) and characteristics (rows); colors are the same in Table \ref{tab:entropy_all}. Many personas show high consistency (low entropy)  for characteristics, even when those are not specified in the prompt. }
    \label{tab:combo_exp}
    \vspace{-10pt}
    \end{table}

\end{document}